# MSc in Computer Science

## MSc Project Final Report

### AI BASED CHATBOT: AN APPROACH OF UTILIZING ON CUSTOMER SERVICE ASSISTANCE

**By:**

**Rejwan Bin Sulaiman**

**[1436184]**

rejwan.binsulaiman@study.beds.ac.uk

**Supervisor:**

Dr Paul Sant

**Department of Computer Science and Technology**

**Faculty of Creative Arts, Technologies & Science**

Submission Date: **15/11/2019**

# Declaration

Following is the declaration of the project:

- I can confirm, the entire work is done solely by me, wherever I have used any existing idea or graph, I acknowledged and adequately referenced.
- The total page count of this thesis paper is approximately 108 pages.
- I allow this thesis paper to use on the Blackboard for other students to gain access from the past dissertation about the related topic unless and until it is entitled as confidential.
- I understand and admit that thesis paper will be uploaded to the Turnitin for plagiarism check where the software will compare the similarity with other sources over the Internet. I admit if the plagiarism is detected over the limit, based on the further investigation, I can face disciplinary action.
- I admit the project is completed at all stages with the serious consideration of ethical values and legal legislation

Signature: *Rejwan Bin Sulaiman*   Date: 15/11/2019

University of Bedfordshire

# ACKNOWLEDGEMENT

**"Oh Lord, that lends me life lend me a heart replete with thankfulness."**

*-William Shakespeare*

First of all, I would like to say special Thanks to Dr Paul Sant for his help and guidance throughout my project. His all-time support and personal attention for the thesis report had helped the project to be in the proficient stage. AK Traders London LTD team had supported and provided with all the things I required to complete the project. My special Thanks to our team manager Mr Masud Rahi, team leader Mr Jamil Khan for their assistance and help at all stages and also, it would be unprofessional if I do not thank Mr Amer Kareem for his support in the time of need. A big thank to all of my friends and family members who are always available to support and encourage throughout the project.

University of Bedfordshire

# Dedication

The dedication of this project goes to my Late Grandfather M. Ibrahim Ali Chatuly, my father, MD Sulaiman Bin Ibrahim Ali, my mother Nasima Akhter and my wife, Zakira Khanom. They have always been a source of encouragement and motivation for me. Without their mental and personal support, this would not have done.

# EXECUTIVE SUMMARY

Providing the best customer experience is one of the primary concerns for the firms that are based online. The advancement of machine learning is revolutionising the company's attitude towards the client through improving the service quality by implementing chatbot solutions, which gives the user instant and satisfactory answers to their enquiries. The acceptance of this technology is increasing with the new improvements and efficiency of the chatbot system.

This thesis paper will cover the concept of chatbot system for the company, i.e. AK Traders London LTD. It involves the research work on various chatbot technologies available and based on research, use them to develop a chatbot system for the company. This system will work based on the text as a conversational agent that can interact with humans by natural language.

The main objective project is to develop the chatbot solution that could comply with complex questions and logical output answers in a well-defined approach. The ultimate goal is to give high-quality results (answers) based on user input (question). For the successful implementation of this project, we have undertaken an in-depth analysis of the various machine learning techniques available and followed well-structured implementation to figure out the best solution for the company. The primary concern of this project includes natural language processing (NLP), machine learning and the vector space model (VSM). The outcome of the project shows the problem-solving technique for the implementation of the chatbot system for the company at a reasonable quality level

University of Bedfordshire

# TABLE OF CONTENT

# FIGURE INDEX

# 1

# INTRODUCTION

**CHAPTER**

## INFORMATION IN THIS CHAPTER

- ❖ OVERVIEW
- ❖ CHATBOT: DEFINITION
- ❖ CHATBOT: BRIEF HISTORY
- ❖ MOTIVATION
- ❖ PROJECT SCOPE
- ❖ HYPOTHESIS AND RESEARCH QUESTIONS
- ❖ AIMS AND OBJECTIVES
- ❖ ACHIEVEMENTS

# 1.1 OVERVIEW

Lately, several firms are moving towards the machine learning technologies to deal with their clients. This trend is common in companies who deal with big data, unstructured data and text. This includes those who specialise in marketing and businesses. According to the research, the amount of unstructured data is growing at an exponential rate.[4] Therefore, the primary concern is to have logical control of this kind of data, which is significant and requires development and improvement at all stages.

Machine learning technology is one of the booming technologies which is still under the process of development. A chatbot (question-answer system) is one of the primary application systems that work using machine learning. It has been a while; this system is in us. However, its operation is mostly based on factual figures. There are available techniques that are based on the NLP and Information Retrieval (IR) that are regulated with the statistical machine-translation (SMT). NLP is primarily concerned with giving answers based on the data sets, and it has limitations of exploring the connections of the words. However, IR deals with the generation of new words and sentences that are based on previous replies. It is quite intelligent in the sense that by utilising machine learning technique, it has given out some logical answers.[5]

In this project work, we have created a chatbot for the company which is capable of giving logical outputs by using machine learning techniques. As the company requires an intelligent system to deal with their client, therefore we have focussed on the chatbot in terms of making it efficient such that it can even reply complex questions with the sensible answers. This project can be significant for many other firms that require an intelligent system to perform the best customer service. Chatbot system could be advantageous if its' operation is integrated with social media and produce prompt reply while dealing with the number of products without the involvement of human agents. This project is focussed more on the number of techniques while designing the chatbot system by using machine learning technology.

## 1.2 CHATBOT: DEFINITION

In simple words, a chatbot system can be defined as the program, that deals with the simulation of conversations with the human user by using the platform of the Internet. It is a kind of machine-based human-like an agent that is available at all time to process the enquiries. The operation of this chatbot system works based on the fact that it can understand the human enquires (mostly in the form of text) and produce the corresponding output. The history of a chatbot is as old as the history of computer science itself. It is understandable by the simple test performed by Alan Turing, one of the experts in late 1950s [1] where it was to find out if the person is communicating with is human or computer without knowing. This test has the great feature of making the system perfect in the sense that it becomes impossible to differentiate between the human and machines. In the real World, chatbot system is still at an early stage of achieving that efficiency where it could be possible to chat about any topic has been predicted n 1950s. This trend of chatbot having capability to understand any topic can lead to conversation flow until the ultimate target is achieved. Experts and researchers have given efforts and concerned to achieve efficiency by adopting numerous behaviour trends. For this project, the consideration of chatbot system is the communication via text between the human and the computer program to process the enquiries and gives the logical output.[6]

## 1.3 CHATBOT: A BRIEF HISTORY

The origin of chatbot goes back to 1966 when a program known as the ELIZA,[1] was introduced with the capability to re-phrased input given by the user (which we called it as the natural language processing NLP). It was indeed one of the simplest forms of the chatbot with the feature of processing user input, and afterwards, it has captured massive attention by the researchers and scientists. Furthermore, this system successfully able to fool many people.

After ELIZA is introduced, the concept has been used for several decades with some minor changes and addition of some more features which typically includes the voice recognition as well as the understanding the emotional feelings in various ways. Later by 2001, the new trend of the chatbot was introduced known as the Smart-Child which was operated in the MSN and AOL messengers.

Afterwards, by 2006, the significant advancements and improvements were introduced in the chatbot system, which was developed by IBM.[3] It was typically introduced to make it run to win one of the famous TV show 'Jeopardy'. This strategy is followed by the use of the advanced concept of NLP, where it involves the instant retrieval of the information. However, the major drawback of this followed up with a long conversation only, that means it could not continue to have a two-way conversation with any user. [25].

Ultimately, by 2010, the most prominent concept of the chatbot was introduced, which is operated by virtual assistants like Cortana, Siri, Alexa. The introduction of these virtual agents has brought a great revolution in the World of chatbot system. These conversational agents are involved in the two-way conversation with the logical dialogue concept with human users. During the same time, another significant improvement was introduced by Facebook while introducing the Messenger that is capable of creating the conversational agent for other firms that are not based on AI. As it can be understanding, that significant development was done during the early development phase of the chatbot system. The system is still developing and improving while many firms are adopting this technology.

## 1.4 MOTIVATION

Many researchers have been working in this field since the introduction of chatbot to make the customer service experience more productive and flawless.

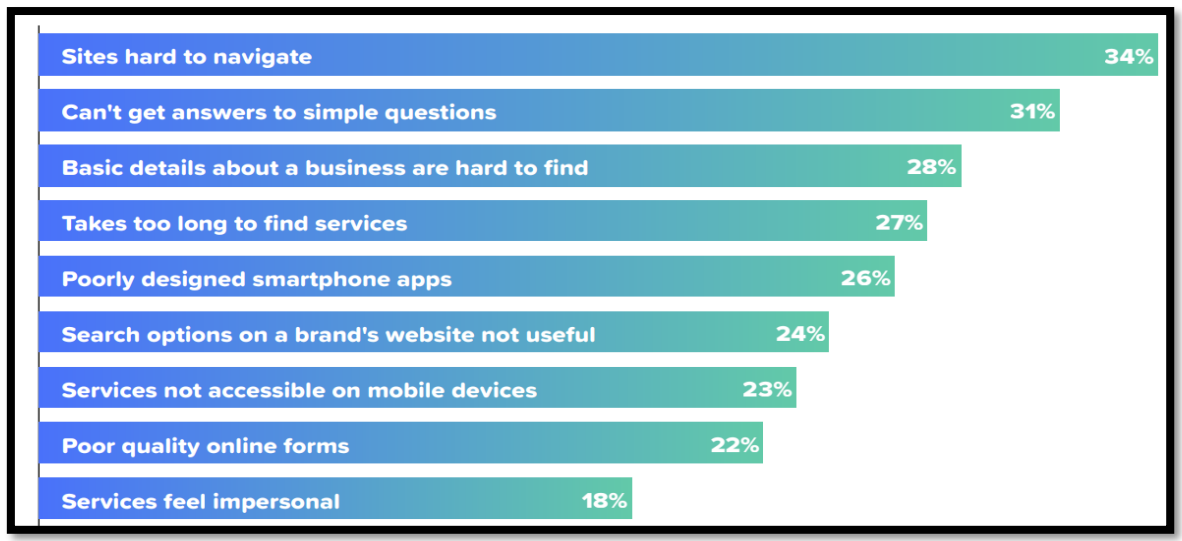

*Figure 1:Customer Service Experience*

the following observations have been made after taking the survey into account:

- Most of the users facing common online communication-channel issues. 34 % are saying, the website is harder to use (navigate), and in contrast, 31 % of users are claiming that they are unable to get answers to their simple question with these communication platforms.
- On the other hand, users are facing some advantages of this system as well, i.e. 24/7 service as claimed by 64 % audience, similarly quick reply claimed by 55 % and the reply of simple question 55 %.

The above result from the survey done shows that the expected outcomes are not achieved by using this ordinary conversational channel. It shows, the use of chatbot is not as useful to some alternative solutions. Currently, the efficiency of a chatbot is still at its early stage, which requires significant improvements and developments. Around 53 % of the users from the survey claimed that the chatbot system is 'ineffective or 'rarely effective'. It shows the potential advantages of this technology that can transform life concerning the real implementation done so far in this field. The youngers are more concerned about the positive impact of this system.

After the survey, our primary motivation is to use this system for the potential benefit for the company. The use of advanced techniques of NLP can certainly cause improvement. As our company is primarily concerned about the best possible solution to use this system for the best experience of their client, therefore we are putting our primary focus to use it effectively and try to minimise all those parameters that are causing problems at users end.

## 1.5 PROJECT SCOPE

This project is designed for a company who is willing to provide the client with the 24 hours online virtual agent to solve their concerns and enquiries. The achievement of this project is to be used by the company to implement it on their official website and make it available for the clients and the public. As this is followed up with the emerging technology, therefore, there is less amount of information that is available. The future impact of this project is uncertain as the system is new and under development and improvement. The project is designed based on recent research and available technologies; however, this tends to improve and update in the future for the best interest of the clients.

## 1.6 HYPOTHESIS ANS RESEARCH QUESTIONS

The primary consideration for the chatbot system in the company is to decrease the lacking's of communication gaps between the company and its clients. The chatbot is here to solve this issue. By implementing a chatbot system in the real world could make the efficiency of customer service by making an instant decision on the human input questions.

The proposed chatbot system for the company can significantly improve the customer experience by answering the questions, for instance, "What is the price of Heathrow from E6?", "Where is my driver I have called for service?". So, answers to these questions by virtual agent can significantly reduce the time and stress of the customers by instant replies. The primary focus of this project is to enhance customer experience towards the company by successful implantation of the chatbot system, which is available at any time of the day for assistance. In implementation phase of the chatbot system, the following questions have been taken into account:

- How can the chatbot system be efficient for the company by reducing the runtime?
- How can the chatbot system be user-friendly?
- How chatbot output generated is efficient and reliable to the user input?
- How can a chatbot be considered as an intelligent system for information processing?

While studying the above research questions, it is significant to understand the customer experience towards this new system. User acceptance, reliability and intelligence are also the factors that we have looked into. So, it works based on the fact of how this system can be advantageous for users concerning the already existing functionalised or information available on the website. On the other hand, it is also necessary to perceive the information provided by the chatbot system is valid and factual. It is the ultimate goal to establish a system which can be accepted and compensate with the lack of techniques available at present.

## 1.7 AIMS AND OBJECTIVES

The main aim of this project is to improve the customer experience by introducing this new way of communication between the customers. The objective is to build a chatbot system on the company's website that is capable of processing customer enquiries. It includes even complex questions and gives several replies in a transparent manner. We aim to give replies to which sensible and logical answers can be given. The proposed chatbot system will use the machine learning techniques to generate the outcomes and retrieve the answers from the database.

In the first part of the project, we have focussed on the various research techniques available for the chatbot system. Afterwards, methods are decided that is capable of processing complex questions while giving definite answers.

## 1.8 ACHIEVEMENTS

Following are some of the significant achievements of this project, that we aimed to gain:

- ➢ Design, develop and implement a user-friendly chatbot which has been improved by time with the user's experience.
- ➢ Evaluation of this chatbot system, So, this system could be used and adopted by potentials customers.
- ➢ Complete research work and learn more on the selection of the best techniques for the natural language processing and other various tools that can be used of the implementation of the chatbot system.
- ➢ Learn in-depth about different machine learning techniques and algorithms.

# 1.9 REPORT STRUCTURE

The whole report is structured as follows:

**Chapter 1, We** have discussed the introduction of the report. We also included definition, history of the chatbot. Moreover, we also included the motivation, scope, aims and objectives of the report.

**Chapter 2** is all literature review, where we have discussed various available technologies and compared them and selected best based on our requirements. In the last sections, we explained NLP techniques, the vector space model and machine learning models that can be applied in chatbots.

**Chapter 3** includes methodology, where we have discussed the method selected for the development of the chatbot system.

**Chapter 4** is the analysis part, where we have analysed the data we have collected from various source as required from our data collection methodology.

**Chapter 5**Designing and artefact development is also part of the methodology. Here we have illustrated the different graphs, diagrams related to the chatbot.

**Chapter 6** constitute the implementation and testing of the chatbot system based on designing while using machine learning techniques. All the steps for the development of the chatbot system are discussed here. Error checking and evaluation of the implemented chatbot system also included in this chapter.

**Chapter 7** includes the conclusion of the whole project work while indicating the specific limitation of the system with future improvement suggestions.

# 2
# LITERATURE REVIEW
CHAPTER

**INFORMATION IN THIS CHAPTER**

- ❖ Overview
- ❖ Chatbot system
- ❖ Chatbot technology
- ❖ Types of chatbot
- ❖ Comparison of chatbots
- ❖ Functions of chatbot
- ❖ Syntactic analysis
- ❖ Vector space model
- ❖ Machine learning models
- ❖ Support vector machines (SVM)
- ❖ Neural networks
- ❖ Critical evaluation tools techniques and technology

## 2.1 OVERVIEW OF LITERATURE REVIEW

The significant contribution for the company from my project placement involves the introduction of 'chatbot'functionality within the company's website, which is the cause of unprecedented change in the user experience. While implementing this system, we have gone through the various techniques and tools that been used previously, and still, are in use in the development of many systems.

If we try to analyse the literature review, numerous techniques are available to implement a chatbot, and they differ based on various factors and functionalities.[4] For the development of this project for the company, we have considered a variety of tools and technologies used for chatbot system. For example, various NLP methods, Vector space models and tools have been compared with a variety of models that used in the machine learning process for research to implement this system.

## 2.2 CHATBOT SYSTEM

In simple terms, a chatbot is a program that can be used for one to one conversation with the users (humans). The program is specialised to create the efficient interfacing of inputs that are received from the user, and while using this input, it processes using various methodologies (e.g. machine learning) in order to provide output to the user. The primary specialisation of a chatbot system is to ensure the user, can have a meaningful conversation with another person [4].

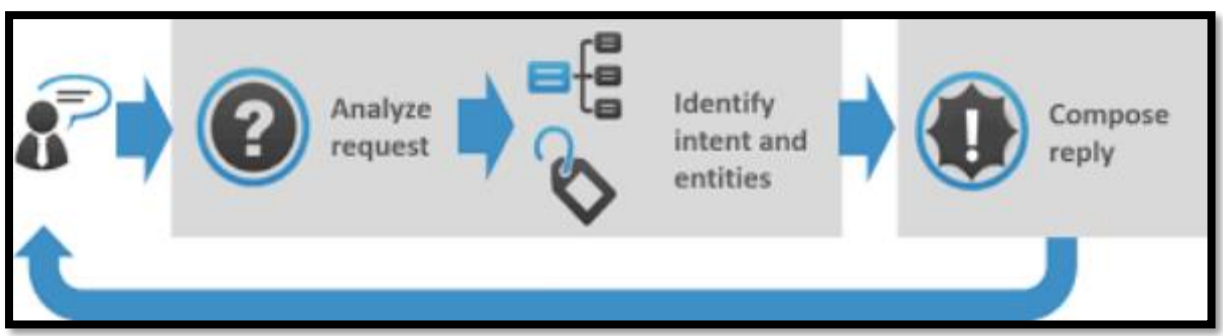

*Figure 2: General Flow of Chatbot System*

The existence of this kind of system relies on various factors that include key-word matching, the similarity in numeric or string and other techniques of processing of natural language. There are several chatbots which are implemented are complex enough to understand user input. The chatbot is commonly used in many web applications which are involved in providing assistance to the user while providing them with a reply to their enquiries.

Nowadays, there is quite several technologies and types of a chatbot that are in use which is differentiated in many ways, including functionality. The chatbot is available in a variety of forms.

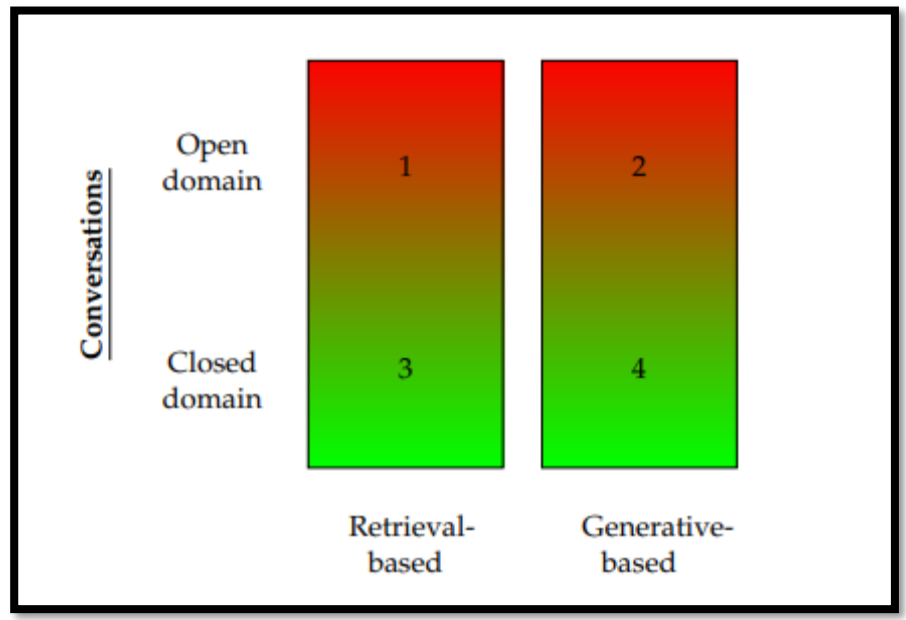

*Figure 3: Type of Chatbot Based on Conversation [16]*

The above figure shows the main categories of a chatbot. In contrast to a closed domain, the open domain is entirely free to cover a variety of subjects to process, whereas a closed domain is restricted to specific areas. There is nothing very different from its operation. To understand this, we can take the example of Twitter status, which can be considered as the open domain; however, ordering pizza online suggests following a closed domain. On top of that, enquires related to finance come between open and closed domains [26].

The differentiation of chatbots can be either retrieval or generative messaging. Retrieval is referred to as the use of information from the available data. However, a retrieval-based chatbot follows the sequence of an enhanced matching process of string and sentences, and it can also involve machine learning tools. These terminologies follow the use of pre-defined techniques for the output process of any given input. Generative chatbots are more complicated in terms of generating their response rather than using pre-defined material [3].

The following are different types of chatbots that are differentiated based on the responses they provide:

## 2.2.1 OPEN-DOMAIN (RETRIEVAL BASED RESPONSE)

Those outputs which are based on retrieval rely on the fixed set of data, and this lead allows any possible enquiry to be dealt with. This sort of scenario cannot be implemented in the chatbot. Because the fixed set of retrieval is hypothetical set that should be made with, any possible question a human could think of, obviously that cannot be done, and hence this type of chatting bots is not possible to create. [26]

## 2.2.2 OPEN-DOMAIN (GENERIC BASED RESPONSE)

It is also concerned with the generation of a valid response towards any possible enquiry. The solution for solving these enquiries is based on the generic response, generally referred to as the AGI, which is the abbreviation of Artificial General Intelligence. It can be understood that the chatbot can carry on with similar tasks which are based on intellect as Human does. There is still an open research work going on in this sector.[26]

## 2.2.3 CLOSED-DOMAIN (RETRIEVAL BASED RESPONSE)

This type of chatbot comprises the use of individual datasets along with the tex, which is based on the specified domains. So, any enquiry generated is handled with the output from those datasets. Anything, which is not available in the domain is not facilitated as it does not provide a solution to the enquires that are not available in domain dataset. So, while using this technique of chatbot, most organisations use this, in case of any enquiry which is not available in a domain is referred back to the human for further assistance.

## 2.2.4 CLOSED-DOMAIN (GENERATIVE BASED RESPONSE)

In this type of chatbot, there is the use of intelligent machine technique to give out the solution to the enquiry. The output that is produced is based on the datasets. However, it acts as the human, in terms of providing its suggestions by giving advice and a solution to the enquiry. The drawback of this kind of chatbot is the use of big training data, and it causes an increase in the level of the problem while generating the outputs that contain significant grammatical errors.

## 2.3 COMPARISON OF CHATBOTS

While contrasting the various chatbots explained above, there are several companies, and they look up for their requirement in the process of chatbot selection. Open-domain retrieval-based response is not possible to implement so far. While in various trials performed by Google and Microsoft, there is nothing been achieved that can make this chatbot successful. So, most of the time, closed-domain is usually used in the selection process; for example, financial domains.[21] Retrieval based as well as generative based closed-domain are commonly used by many firms. However, there is still work going on to improve the efficiency of these chatbots, and it certainly involves the expansion of the chatbot domains. The priority always leads to the best possible accuracy on of grammar and others; therefore, to compensative with this requirement, retrieval based chatbot is quite apparent. Shortly, it is predicted the use of generative based response when its performance and efficiency will be enhanced in specific ways of having big datasets for the solutions of maximum answers [21].

## 2.4 ADVANCED CHATBOT SYSTEMS

The chatbot is always considered to replace the humans by chatbot if that is providing the best response. We can think as it is an online friend that is always available to make some talk and conversation. In simple words, it is more like a match patterning system which is expertly made to act like humans as much as possible. Nowadays this match patterning system is mostly considered for any automation and other chat channels as well. There is an automated-online-assistant (AOA) is one of the systems that use the same concept of the chatbot, and it allows the users to chat and carry on with some simple tasks. Chatbots commonly utilise straightforward match pattern concerning the provided input, however in contrast to those AOA utilise more advanced methods which includes the use of NLP, named-entity recognition and analysis which is based on both semantic and non-semantic. These types of techniques can also be used in question-answers as well; however, these are simpler as they are involved in only answering the question but not to perform the task. This can also be explained by the logic of QA as this system is only involved in answering questions, not chat itself as a whole [6].

## 2.5 FUNCTIONS OF CHATBOT

There is a broad spectrum of chatbot functionalities which are very vital for many applications. The real need is to use this system to able to gather information from the user and can chat rather than just simply giving answers to the users. Following are some of the methods of chatbots which are based on their functionalities and limitation of its operation:

## 2.5.1 ENHANCING THE CHANCE OF RIGHT ANSWERS

The fundamental requirement for the companies to adopt chatbot for their clients to benefit from this system by having correct answers towards the enquiries. It is used to ensure the maximum probability of giving out the right answers:

- There is a variety of algorithms that are used in machine learning which is capable of giving out the exact percentage of the chances of getting the correct answers towards the provided input. Following graph, figures show the binary classification of probability in amazon machine learning [7].

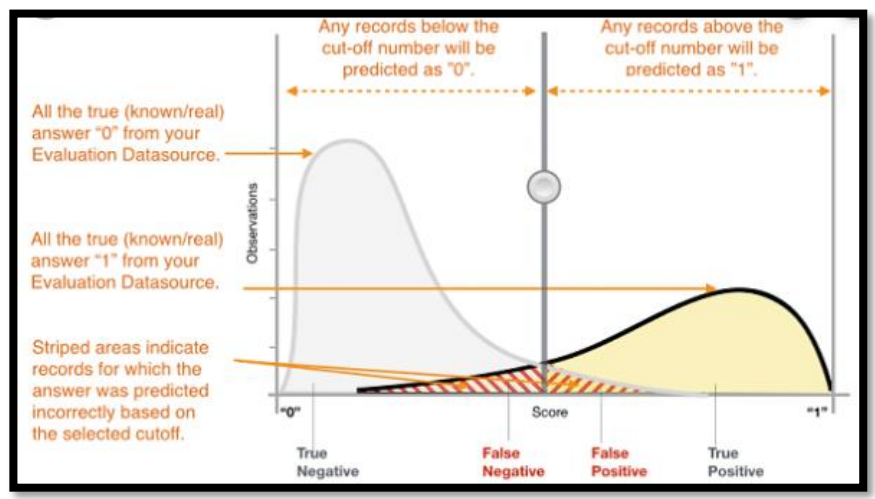

*Figure 4:Binary classification of Probability [7]*

So, while keeping this in mind, the chatbot can be designed as the way it could output an answer only if the probability (percentage) is high. However, if the certainty is lower than program can follow to send the enquiry to the expert and from there it looks upon the filled form of all enquiries, output follows up if the expert is agreed with the answer, and it only required click and sent, in case of any disagreement, it has right to modify or even format it.

- The simple method of chatbot operation follows up with the dataset which has allocated answers and questions produced by the expert, so once a new question is received, it looks up with the similar allocated question in the dataset and gives the reply. For more certainty, chatbot follows up the procedure of asking question to the user "does your enquiry/question matches with (dataset)?", in reply to this question, if the received input is positive, chatbot will give the right answer, similarly if the reply is negative, it goes back to the expert for further processing. [8].

- In the case of QA, there are always a variety of questions and answers; however, in reality, it follows up with a similar kind of questions (same). So, in this method, it can easily follow up with the set of standardises questions and answers. Therefore, it can be classified as the one question which can be input in a variety of ways. Some dataset is followed up with the higher chance of giving the right answers from a set of the same class of variety of questions. [7]

## 2.6 TECHNIQUES OF ANSWERING

There are a variety of techniques that are in use which is based on the AI system. Information-retrieval is a kind of approach which is used for solving the query. It is originally based on the accessing data, and it uses to retrieve the list of possible approach towards the enquiry instead of just giving out the answer. Hence, it gives out to the user, and in this way, they can track around the information given and find out the exact solution to their enquires. This can be easily understood with the example of the Google search engine. It follows up with the user to ask a question in simple words, and it retrieves the list of various small answers towards the question [9].

In our case, it is essential to give out one specific answer, it might contain a link to the other websites; however, it will not retrieve the information which is based on the list of documents. Technically, we set out the best answer based on the highest probability of certainty, which is followed with the list of answers to the questions.

### 2.6.1 QUESTION REPLY

Sometimes, an additional question from the user is required as it follows up with the complex question that needs some more detail from the user to process it. e.g. if a user asks at when he is eligible to get a pension, then the system will regenerate a question asking the date of birth of that person. It is an excellent feature of the chatbot system while replying with the new question as it helps to get the exact information about the user that ultimately help to get the right output.

It is not an easy task to generate a reply question while keeping the relevancy of the context of the enquiry, as this requires from the client the sort of information they precisely needed and inform what information is missing. So, to make this sort of design in chatbot system, it is required to use of template that can be helpful in terms of getting information such that if that question is linked with most specified frequently used questions, then the program will look up for the year of birth mentioned and in case if it is not there the program will going to reply a question asking for the date of birth. This kind of system doesn.t provide flexibility, and it can only be used in limited and specified cases. Currently, there is no technology been developed to solve this in some alternative way [10].

## 2.6.2 NATURAL LANGUAGE PROCESSING (NLP)

It is a process which let the machine to learn and understand the human language. It utilises the hierarchical procedure for the processing of natural language. This kind of language processing is considered to be complicated in the IT world; reason been is the complexity and ambiguity of the natural language. This system works bearing in mind that if the computer needs to understand the meaning of the natural language, it does not only involve the meaning of single words or sentence rather than it has to follow up with the understanding of the whole concept. NLP is divided into the following components [11]:

### 2.6.2.1 MORPHOLOGICAL ANALYSIS

This type of analysis involves the use of understanding the smaller elements found in the word. In the first level of natural language processing, it only looks up for the words that are taken from the text, and at this stage, it does not contemplate the punctuation of the whole text. Moreover, these are further narrowed and analysed into smaller components. The following figure shows the simple illustration of the algorithm used in the morphological analysis:

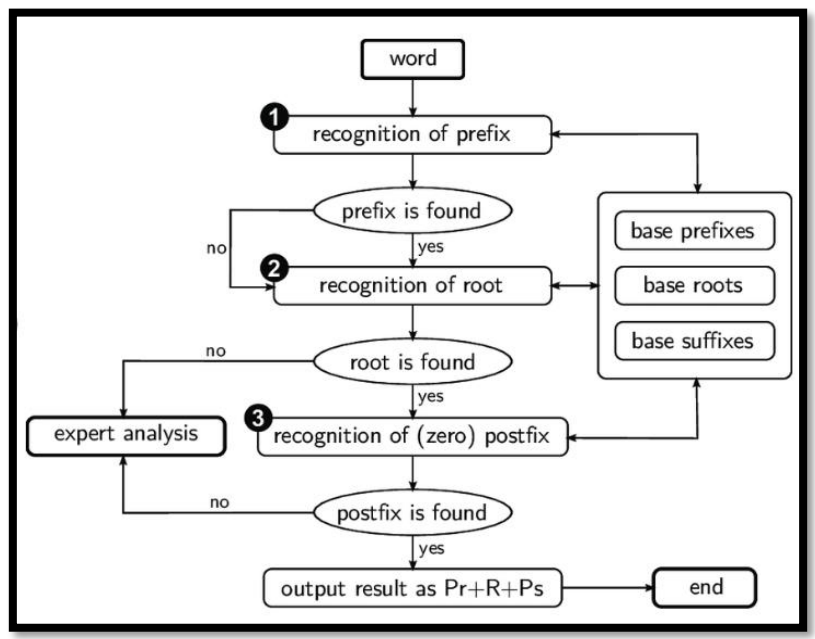

*Figure 5: Morphological Flow [49]*

Following are some of the essential terms that are used in the morphological analysis:

### 2.6.2.2 TOKENIZATION

It is techniques which are used for the process of breaking down the whole text into different words, sentence and other smaller components which are referred to as the tokens. Those tokens which are processed is taken for further processing as an input. Furthermore, at this stage, tokeniser is required to follow up with the removal process or transformation of the abbreviation, punctuations and other white spaces. The significant involvement of the tokenisation process is the identification of some useful or meaningful words in the whole text. [49]

### 2.6.2.3 REMOVAL OF STOP WORD

These are commonly used words in natural language, i.e. 'a', 'of', 'the'etc. Their words are not used for the separation of the context; instead, these are just used for joining the sentences. In case, if these words are not believed essential in the text, then these words can be removed as per domain. Removal of these words is quite a difficult task as it involves in-consistency and ambiguity to differentiate between the meaningful or meaningless presence; therefore, for the generation of the stop-word list is not an easy task. It is essential to perform the removal process of these stop-words as this will lead to the small data-size, and ultimately, it causes better performance in the classification of the whole text.[24]

### 2.6.2.4 STEMMING PROCESS

It is the process that can be used for the reduction of words that are conjugated. Moreover, it causes the words to come back to its original form. It can be understood by taking the word, "speak", "spoke"and "spoken"; these three words are classified into one word which is spoken. This method causes the ease of matching the text with the same context. [4]

### 2.6.2.5 AUTO-QUERY EXPANSION

It involves the use of some additional words with the original query, and this will lead the matching process of the query among the text. Some of the conventional methods that are used for auto-query expansion is the use of synonyms, stemming process and spelling-errors fixing.

**2.6.2.6 PART OF SPEECH (POS) TAG**

It is the process of tagging word in the text concerning the PoS. It is understood with the example of thinking about tagging a word with the part of speech (PoS). This process can be done either by a rule-based approach or with the statistical approach. The approach of using this PoS based tagging is quite useful in the generation of the output in the sense of parsing of the sentence, and this causes the less ambiguous of the words used in NLP.

## 2.6.3 Syntactic analysis

It is a type of analysis which concentrates on the components in the sentences. The second stage under the NLP explains the transformations of words into structures which can show how the sequences can have a relation among them.

There are some terms which are vital for the syntactic analysis. These are discussed in the following:

**2.6.3.1 PARSING**

It is an operation which transforms the sequence of words into a formal structure based on grammar. Parsing delivers a tree that explains the relationship between the input and the words in case of computational linguistics. Deep parsing is responsible for the complete tree building, whereas shallow parsing only builds only a portion of the tree for a single sentence [12] [1]

**2.6.3.2 BAG-OF-WORDS**

This model represents the texts in a straightforward manner. A text is displayed like a group of words where the words have no relation between each other. Besides, no grammar is acquainted with the texts. This model can only display the words and the multiplicity that they have, and this model has a usage in classifying the documents. [12]

**2.6.3.3 N-GRAMS**

The model, N-gram, is widespread based on storing spatial information, whereas the bag-of-words do not provide much of texts. it has been stated that this model is widely utilised to stem. By using the model, it is possible to anticipate the usage of a word in a sequence. The total amount of words that have been regarded can be denoted by N in this model.[13]

For example, if N=2, the sentence transformation may look like, "Where do you want to go?"into different segments like "Where do", "do you", "you want", "want to", "to go". This anticipation of the recurrence of particular words can be resulted due to the frequency of the words occurred. However, in pragmatic practice, it is unlikely to have the word combination present in the data set. This problem can have a solution provided that the probability distribution is undertaken for the appeared word [14].

### 2.6.3.4 NAMED ENTITY RECOGNITION (NER)

NER is responsible for the arrangements of proper names such as the names for people or even places. There are a lot of researches going on in this area. The current approach of this report would be to trigger and find the pattern to match with an option to look up. In the case of lookup, the system can only be entitled with the capability to have recognition of the stored list. Also, it is a fast approach that has the adaptability to another text. However, there are fixed costs of maintenance and collection of the entities to provide definite problems regarding the variation among names. For instance, "Mount Everest "can be considered which can be recognized as "Mount + Capitalized Word". In the case of pattern matching, the patterns are constructed manually. For instance, "<Name> lives in <Location> on <Date>", will yield the result of "Jessy Lives in Boston for 3 years" [2].

### 2.6.3.5 WORD SENSE DISAMBIGUATION

This is utilized in order to determine the meaning of words which have been used in texts. It is known that one word can have a difference in meaning, and it is one of the difficult tasks to differentiate the words. The associated methods in the case for disambiguation can have inclusion of knowledge-based methods, empirical methods, and AI methods [16].

### 2.6.3.6 LATENT SEMANTIC ANALYSIS (LSA)

LSA is a methodological process which has embedded assumption that the words which have identical meaning will be most likely to appear in the texts. The terms, as well as the documents set, are analysed in this LSA relationship. In order to contain the counts of the words, a word-document is formulated where the text documents are kept in the columns, and the unique words are kept in rows. Similarities are kept intact in between the texts where the reduction of unique words is being made by singular value decomposition.[27][28]

## 2.6.3.7 SEMANTIC (ROLE) LABELLING

The semantic labelling identifies the acts of the words in texts. An example can be given in this regard such as, "Jessy lives in Boston". The semantic labelling will be able to identify "Jessy"as the person who lives, the verb "lives in"as well as "Boston" which is the habitat of the person. It enables the reader to understand the meaning [39]. [41]

## 2.6.3.8 LEXICAL ANSWER TYPE (LAT)

LAT is responsible for indicating the entity [23]. An example can be given to understand it better such as, "Who is the president of USA?"and here, LAT is the "Who" since it identifies the person who will be responsible.

## 2.6.3.9 RELATION DETECTION

Relation detection is utilised in order to discover the relationships between syntactic and semantic in sentences [23]. An example can be given like, "Who is the president of the USA?", the relation can be found the ad (president, ?x, the USA)

## 2.7 VECTOR SPACE MODEL

The vector space model was initially utilised in the SMART Information Retrieval System [30]. This mathematical model can be utilised in determining the vectors to represent the textual documents. The comparison can be made by having this model; hence, the calculation can be done to find out the same things in-between the documents.

This vector space model creates a matrix, and then the value assignment is ensured, which will be compared after the words make an appearance in text documents. One of the most famous weighting operations is called the TF-IDF weighting system, which has already been explained in full in the prior section of this project report. The total number of words is the main focal point to determine the dimensionality of the vector, such as the total number of words used in the documents and the words or the documents which will be compared.[26]

In order to make a comparison between the text documents, it is a must to undertake vector operations. The operations can be called as the measures which can be done to find similarities. The similarity measure operation can provide a ranking which can depict the relevance and then the comparison will take place between the text documents. The similarity can be measured upon considering a formula which goes by the name 'cosine similarity'. The formula is given below:

$$\cos\theta = \frac{d_1 \cdot d_2}{||d_1|| \cdot ||d_2||}.$$

Concerning the mentioned formula, $d_1 \cdot d_2$ denotes the intersection of two-document vectors whereas $||d_1|| \cdot ||d_2||$ denotes the length of the two document vectors to have normalisation in the scorecard. As high the value is for cosine, the high similarity is found between the documents. It is imperative to have the normalisation since it can provide surety of having fewer chances for a more extended version of documents in terms of matching is concerned.

One of the limitations of the model is that the document is expected to have a word-to-word match in terms of words with another document. However, by utilising various synonyms and pre-processing techniques, this problem can be got rid of. [18]

## 2.8 MACHINE LEARNING MODELS

Machine learning is a field under computer science where computers usually possess the learning ability without being programmed to do so. The algorithms of machine learning make the data-driven anticipation efficient, and it also allows the computer to get rid of the necessity of learning rigorous programming instructions.

There are different machine learning methods which can be folded into different classifications in numerous ways. Supervised and unsupervised learning can be two ways to the classification. In case of supervised learning, the computer, along with the anticipated variables, can be presented with a view to the usage of the variables. The intention of this learning is to create a pattern which can have alignment with the anticipated values in the results. On the other hand, the unsupervised learning possesses no label, means it does not have any variable. A dataset can be observed with different observations for numerous sets of variables that have observations. The algorithm is responsible for the data structure. This type of learning structure can be used in order to discover whether there are any hidden patterns inside the data or not. [19]

The type of the total output machine learning generates- can be the ways of classification. The classification can provide the answers in multiple classes. The model has been used in the supervised learning model, which can indicate the new class that it belongs to. The regression method can provide supervised solutions and is unable to provide categorical outputs since it only produces continuous output. Clustering method can classify the aggregate data to groups even if it does not have any familiarity with the groups.

This section will discuss the techniques of machine learning, which can be used for chatbots. Also, emerging techniques will be discussed here. Though there are many techniques associated with machine learning, question-answering systems have been taken in this regard. [19]

### 2.8.1 K-MEANS CLUSTERING

K-means depicts the classification for the n types of observations to k groups. The notion of this idea promotes the observation which belongs to the closest mean.[24] The algorithm has functionality that can be divided into five categories:

- ✓ Choose k: the number of clusters
- ✓ Initialisation of centroids
- ✓ Assignment of points to the cluster along with the centroids
- ✓ Re-arrangement of centroids based on the formulated clusters
- ✓ Re-assignment of points to the cluster along with the centroids

The last two steps should be repeated until they do not have a fixed position, and the data points should also possess fixed positions and should not pass to the nearest cluster.

The initialisation of the centroids can be completed upon random choice. However, it is advised that it is essential to utilise the heuristic which locates in the farthest. In this case, though the first centroid has been selected randomly, the second centroid has been selected from the farthest [20].

### 2.8.2 K-NEAREST NEIGHBOURS (KNN)

The classification and regression can be done under the supervised learning method of the kNN algorithm. In terms of classification, a new point can be depicted to classify the new point which can be denoted by k. The assignment of the neighbours belongs to the k, which is a significant vote to be classified to determine the new point.

Here, the classification is done to two different classes which are different in colours such as class 1 (green) or class 2 (blue). If k=3, it depicts the nearest neighbours close to the red star, it would then be classified into blue or class 2. However, if the attention can be placed in k=5, then the star will be given to green or class 1. This gives the information on the sensitivity of kNN to the structure.

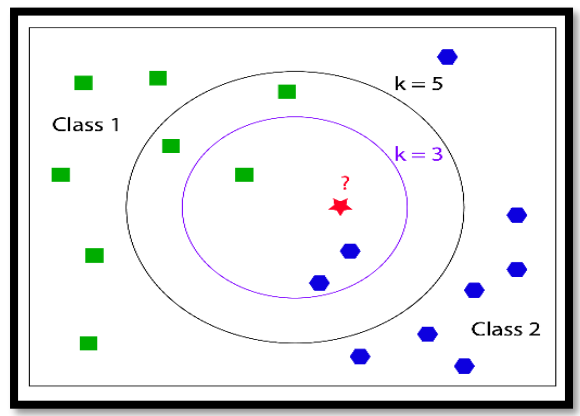

*Figure 6: KNN classification Algorithm [24]*

In case for regression with the kNN input, it consists of k whereas the overall output is to have a property value. The property value is considered as a mean of all the values.

In case for regression as well as classification with KNN, the overall algorithm can have developments if and only if the distance of the neighbours' values can be counted while doing so. Afterwards, the assignment of weights should be done to the nearer and farther neighbours. More weight should be given to the close ones, and less weight should be given to distant ones. Here, d signifies the distance between the neighbour and the new point [21].

### 2.8.3 RANDOM FOREST

One of the productive machine learning models, which is used vastly, is known as the Random Forest that can conduct regression as well as the jobs related to classification. It allocates several weak models to constitute a large and robust model. The smaller models are reflected as the decision trees, and they have the responsibility to split every datum at every possible node.

Several decision trees can be formulated upon utilising Random Forest and its unique conditions as well as features on those nodes. It is supposed to have different nodes in each tree when it is time for selection. Only a handful part of all the features can be provided as regular options. In order to appoint a reasonably new phenomenon to the class, each tree is required to provide a classification which is known as a vote is some cases. This model is designed to select the class that possesses maximum votes of all the trees. [29]

### 2.8.4 XGBOOST

XGBoost is known as eXtreme Gradient Boosting, which is an emerging gradient boosting that has grown interested among the professionals since it is very fast in terms of fast functionality compared to the instant gradient boosting model.[34][32]

Numerous weak rules combine to form boosting and voting is also used to come up with a powerful model such as the Random Forest. An example can be taken with a form of a weak rule where "if there are questions which have five identical words; then they would be identical to each other too". The rules can be discovered by the application of machine learning algorithms. [27]

XGBoost is unique because of the speed it has due to the execution of parallel computing in the comparison. In addition to that, there are different types of benefits which can function effectively against the overfitting. Also, this provides flexibility to the user since the users have the luxury to evaluate different criteria and also can be able to establish the optimisation objective. [21]

### 2.8.5 SUPPORT VECTOR MACHINES (SVM)

SVM is known as a type of binary classifier which is created based on different independent variables. It is a type of learning method that can produce desired outputs. The boundary that SVM can create is called the maximum margin hyperplane. However, to be able to discover the hyperplane, it is imperative to maximise the margin with increasing in the perpendicular distance between the observations of different classes. In the following figure 7, a hyperplane is depicted with different types of small black lines. The bold ones are the ones which are the exact maximum margin hyperplane since it has the potential to create a huge margin compared to the hyperplanes for green or orange ones. In terms of adding a new dataset, it is imperative to know the location of the maximum margin hyperplane. [22][24]

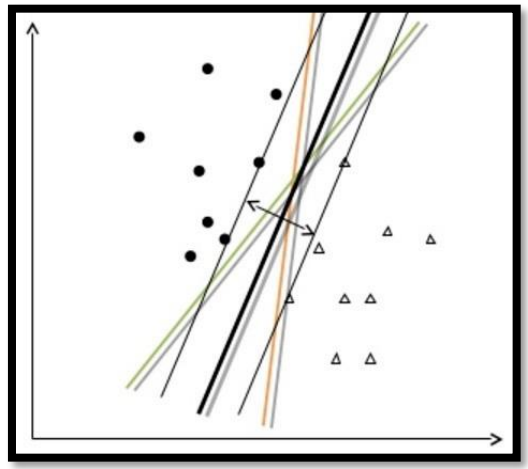

*Figure 7: SVM hyperplane margin [39]*

With the application of linear hyperplane, it gets confusing in some instances to classify the data into two different folds as it can be seen in the figure below [40]. In a higher phase or dimension, this problem can have a solution provided that the SVM is taken into attention. The kernel function is responsible for the calculation of higher dimension. When it is time to translate the linear boundary, the input space turns into non-linear [23][24].

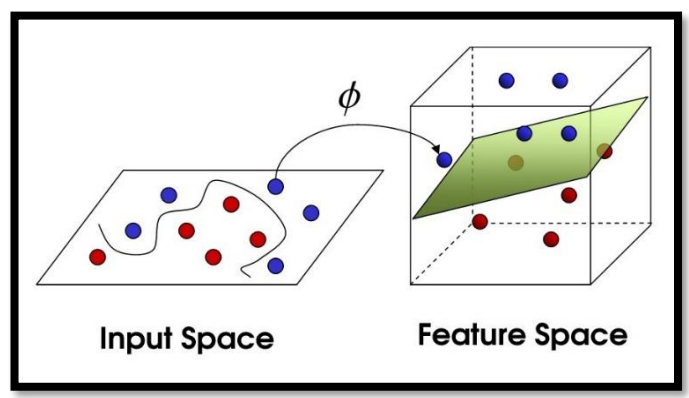

*Figure 8: Non-Linear SVM using kernel [27][33]*

## 2.8.6 NEURAL NETWORKS

The artificial neural network is created on the structure of the human brain. Neurons are amalgamated to each other in human brains, and they compose to cells which are the primary reason for humans to work and think faster. The figure below portrays a network which is called 'artificial neural network'. In case for human brains, a network is located inside the brains where the information is analysed and delivered for further usage. Signals can pass through the neurons and make a connection.[42] Every neuron is responsible for providing a specific function so that the output can be determined according to the input provided. Hidden layers can be found in the artificial neural network, and the layers can be of many numbers which can surpass the numbers of layers in the depicted figure. The input layers can be possessed with numerous neurons, and they will act as independent variables, whereas, in case of the output layer, the neurons can act as the dependent variables. However, the total amount in the hidden layers, as well as the number of hidden layers can only be calculated based on amount and types.[23]

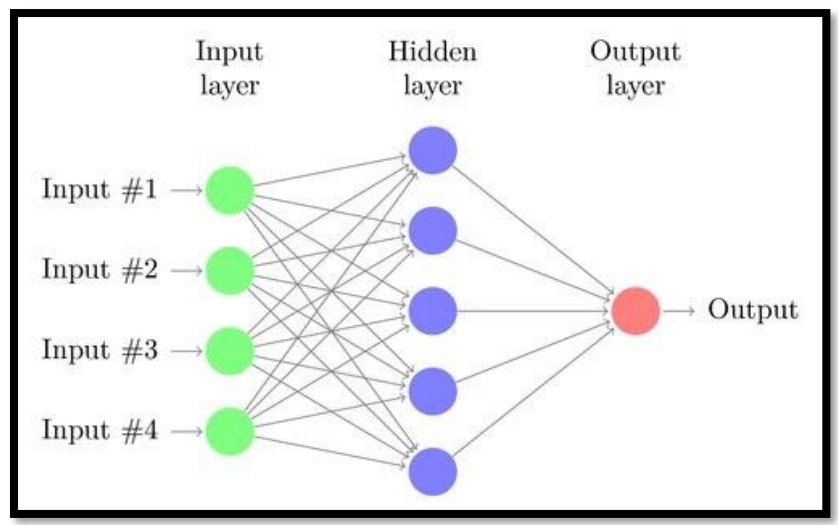

*Figure 9: Artificial Neural Network Layer [22]*

The neural network can be trained with the linkages among different types of neurons in order to destine to the optimum model concerning the training dataset. This is certain that workings need to be done with feedback rest assured, and the network should be capable of understanding between the right and wrong. With all the necessary information it can get, the modification should be done on the connection to have a look at whether the result is okay or not. The alteration on the connection will be put weight into the difference between the outcome based on the actual income and model outcome differ much. The whole process is known as backpropagation in terms of neural network. The functionality of this network begins from the output neuron and moves past through hidden neurons destined to meet input neurons. New responses can be created after the training of the network.[33]

Numerous types of artificial neural networks available. The above figure 9 portrays the Feed-Forwards Neural Network (FNN) since the information can only move towards a definite way. In the chatbot systems, the usage of Recurrent Neural Networks (RNNs) is vastly used since the networks can learn broader contexts which make it unique compared to different types of neural networks. RNNs possesses data propagation which is bi-directional, and the propagation of data can take place from input to output [24].

## 2.9 CRITICAL EVALUATION

As has been discussed in the literature review mentioned above, various tool and technologies can be adopted for the development of the chatbot system on the website. The selection of specific tools and technology follows up with the criteria which deal with the company choice, and of course, the suggestion has been proposed by the IT expert, such that the developed system is sustainable and promising. The above-mentioned technologies give the overall view of the number of possible solutions for the implementation of the system, however, in this section, we will carefully analyse the reason for the selection of tools and techniques that have been adopted in our project.

As mentioned earlier, Natural language processing (NLP) technology uses specific tools and techniques which are ultimately capable of understanding human language, that means it is quite user-friendly in terms of giving the clear command of humans to the machine in a very natural way. This set of rules does not apply to several word-processors. The uniqueness of the NLP technology can be measured by understanding the hierarchical structure of natural language, i.e. letter into word and words into sentences. However, if we take another view, NLP is not that simple as it looks;instead, it has ambiguity, and it is considered as one of the hardest problems in the IT World. The justification lies in the fact that, when a machine (computer) which will be running NLP is required to know/understand not just the meanings of the word or text, it will also have to be taken into believing the actual concept of the whole sentence/words to output accordingly. [31]

Nevertheless, on the other hand, the vector space model can fulfil this drawback, which is based on mathematics, where it displays the word doc in the form of vector. The use of this model which uses vectors for representation can be used to find out the similarity among two documents, and it follows calculations. Initially, this model was beneficial for using it for the retrieval procedure of the data. Thus it was considered as the smart model for retrieval of the information from the database. The vector space model utilises a quite clever technique which is based on the leading of document terms matrix which works in the sense of allocating absolute value to the term only for those which is required to be compared. In order to perform this action, there are multiple ways to do so, i.e. terms-weighting. However, the most common technique for performing this action follows up with TF IDF weighting, as been discussed in the previous section. The exact dimensionalities of the vectors depend on the total words that have been used for the comparison of the multiple documents.[29] To perform the comparison of the various text-documents which are based on vectors, several operations are used. In other words, these kinds of vector-operations are known as similarity-measures. This operating system works in the way that, it gives the ranking criteria that is based on the particular relevance that is followed while comparing and contrasting the text-documents with each other. However, the most commonly used similarity-measure is known as cosine-measure. Moreover, the models, as mentioned above, do not give us the language independence. For our company, we need such a technique that is language independent and can conversate to any language at any time.

Nevertheless, with all this, there is a unique method which covers several drawbacks in the other technologies as mentioned above, i.e. machine learning technology. There is various kind of machine learning techniques that can be used, and they are quite useful in their ways. One of the most popular ways is followed by learning technique, now the concept of differentiation is done based on the type of leaning technique, i.e. supervised learning or unsupervised learning. The working of supervised learning is followed by the presentation of the specific datasets and it is constituted of predictive-variables and result been produced by them. The basic concept of the supervised learning is to make or rule or follow up with the specific patterns or combinations, such that it could be compatible with matching with the result been produced by the variables. However, in contrast to this, unsupervised learning is not followed by predictive variables or output based on them. It is constituting of specific datasets that comprise of the absolute value followed accordingly by the variables which are utilised for the observation purpose. Moreover, it is ultimately following certain algorithms to figure out the order in data. It is one of the prominent types of learning that can look for the hidden structure and order in the data.

While considering the above mention differences, we have carefully analysed the different types of machine learning techniques. Moreover, after taking this into account, we have selected a supervised learning technique for our project.

## 2.9.1 OPERATION OF CHATTERBOT

Machine learning technology which we have adopted for our project utilises the chatterbot. The chatterbot, it is based on the Python library, which is used to give the output based on user input, and it gives out the automated response accordingly. The operation of Chatterbot system is entirely based on the machine learning algorithms that can generate a variety of responses. So, this type of behaviour of this technology creates a friendly environment for the developers to utilise it on the chatbot system for the automated response generation based on the user input.[35]

The processing of chatterbot works by using a variety of machine learning algorithms. The selection of an algorithm depends on the way chatbot is intended to use and the configurations. Following are some of the conventional means of techniques of machine learning technology uses in the chatterbot system.

### 2.9.1.1 SEARCHING ALGORITHM

Performing search is one of the prominent components used in artificial intelligence. There is some technical differentiation between machine learning technology and artificial intelligence. Using the search algorithm is crucial as it is the essential backbone in the process of the chatbot system, which is responsible for performing the instant action based on the user input statement. Following are some of the issue of the chatbot system to generate the response based on the user input:

- The matching of input statements to which is already known by the system.
- The number of times in which matching response is generated.
- The probability of input statements to be known by the system.[36]

### 2.9.1.2 CLASSIFICATION ALGORITHM

There are various logic-adapters which are accompanied by chatterbot that utilises Bayesian classification of the algorithm, which is used to figure out the matching pattern of input statements that can be regulated with the particular set of rules for the generation of the response based on the logic-adapters.

In the field of machine learning technology, Bayes classifier is a part of a probabilistic classifier that uses Bayes theorems. It is considered as one of the simplest models of Bayesian-networks.

To understanding the nature of problem or learning purpose, Bayes classifier is thought to be scalable.

So, the real operation of the Chatterbot system is followed up when the un-trained instance of the Chatterbot is operated or started having no clue of the method of operation. So, whenever a text or statement is used to input to the system, the steps follow up with saving the input in the library and thus process it and produce the Responses accordingly. The operation of ChatterBot is enhanced, or in other words, the accuracy of the responses is improved when the number of received input is increased, i.e. statement or text.[27]

So, in simple words the response is followed by the accuracy based on the amount of received input as it has the capability to understand the generated responses towards the provided input, so when a new input is generated, the program is specialized to look up for the best possible known responses, and in return programs looks for the closest matched statement that has been generated earlier and looks for the best response been generated in that occasion. Therefore, the program carefully analysed the known input responses and based on that, when a new input is received it considers all the knows responses towards the similar input and hence it comes up with the best possible response, thus improving the system response overall.

Following are some of the reasons for choosing the machine learning technique on top of three other technologies mentioned above:

- It makes the system, i.e. chatbot very user-friendly, even the non-technical person can be trained to operate the system by using machine learning technique while this is not the same with other technologies where it requires excellent technical skills and qualifications.

- The use of machine learning can cause ease of conversation update, such that it takes the instant input and process it and give the respective output.

- It also gives the benefit of ease of adding or deleting any topic in contrast to other mentioned technologies

- It allows adding multiple languages, even very different language than English, for instance, Arabic, Hebrew, Bengali. This differentiation makes the system more reliable as the chatbot can be made for any language, any topic and any institution. The capability of ChatterBot of being language independence gives the user a chance to talk in any language of their choice. As the chatterbot operated based on the machine learning technology, therefore the responses generated based on the machine learning can be used to enhance and improve the possibility of response based on the specific knowledge along with the interaction been generated with human and information is gathered

- One of the best advantages of studying this machine learning technology is that it gives the capability of adding any topic of choice without any restrictions. However, this is not the case with other technologies.

- Machine learning technology also has the capability of friendly interfacing with the website as well such that it can easily interactable to the websites, while other mentioned technologies require more complex interfacing to make it work accordingly.

- On top of all, the primary concern of this choice leads to the fact that by utilising this technology make system operation instant and quick with no delays

- As the accuracy level of using this technology is quite higher as compared to the rest of the technologies as mentioned above.

For the machine learning technology, we have analyses carefully various concepts been used, and for the implementation of our Chatbot project for the company, we have used the concept of Chatterbot which is one of the best Python libraries, and it is used frequently in machine learning technology.

# 3 METHODOLOGY

**CHAPTER**

**INFORMATION IN THIS CHAPTER**

- ❖ OVERVIEW
- ❖ ARTEFACT DEVELOPMENT METHODOLOGY
- ❖ WATERFALL MODEL
- ❖ ADVANTAGES AND DISADVANTAGES OF WATERFALL MODEL
- ❖ DATA COLLECTION REQUIREMENTS
- ❖ INTERVIEW
- ❖ QUESTIONNAIRE
- ❖ SURVEY
- ❖ SOFTWARE METHODOLOGY
- ❖ HARDWARE METHODOLOGY
- ❖ FUNCTIONAL REQUIREMENTS
- ❖ NON-FUNCTIONAL REQUIREMENTS

# 3.1 OVERVIEW OF METHODOLOGY

In this section, we will illustrate the design, development of chatbot, and the data collection method. We will also explain different SDLC models, mentioned what models we have chosen to accomplish the project, with the phases of chatbot development. The chatbot is entitled to having a formulation capacity which is done after having a question by the users. Also, the users can have a conversation with the chatbot. The methodology of the development, as well as the execution, will be portrayed in-full, in the following section, which accompanies the machine learning model in the following sector.

### 3.1.1 ARTEFACT DEVELOPMENT METHODOLOGY

There are different types of methods which can be considered with a view to completing the project. In the case of web and software-based projects, there are some models found that can serve the researchers with the utmost benefits since they have the potential and capacity to address the objective of the project. The models are- [38]

- Waterfall Model
- Spiral Method (SDM)
- Iterative and Incremental Method
- V-Shaped Model
- Evolutionary Prototyping Model
- Agile development

### 3.1.2 Waterfall model

Upon giving a thought, it has been concluded that for this project, the waterfall method can possess significant benefits in the project making and the benefits of the waterfall model are given as follows:

- Structured and definite phases
- Orderly management
- Time management from the initiation to the completion point
- Unique deliverables in the System Development Life Cycle (SDLC)
- With the assistance of the sequential model, each phase can be seen with a downward flow until the end of each phase.[37]

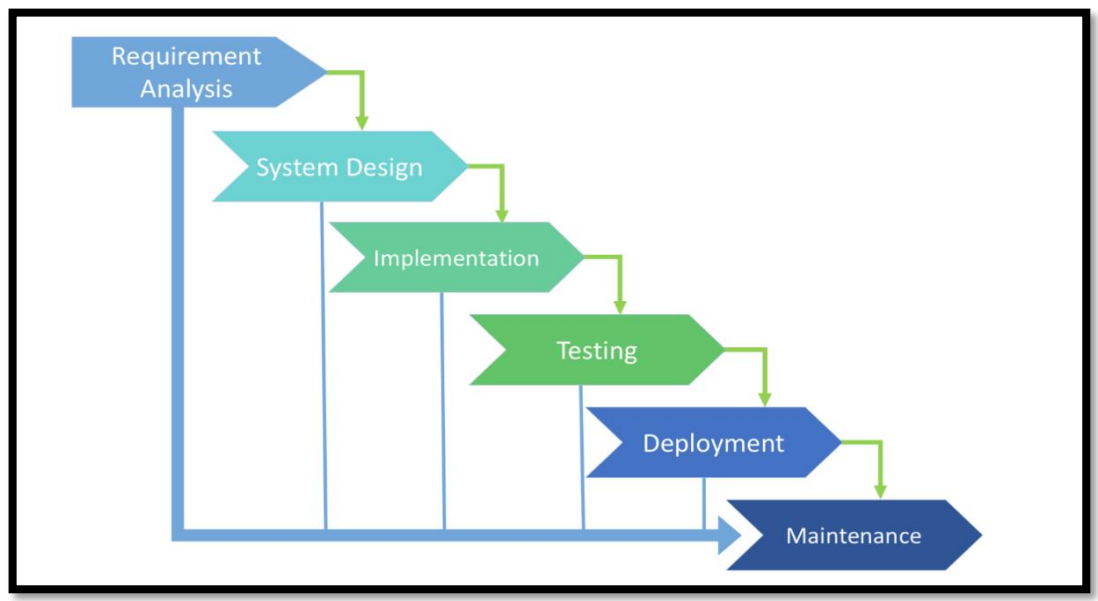

*Figure 10: System Development Life Cycle, Waterfall model*

The figure mentioned above depicts the system development lifecycle, which is initiated with requirement analysis, Gradually and sequentially design, implementation, testing, deployment and maintenance take place. It is essential to make sure that each phase needs to be completed before the following phase can start. Only one phase can go on while other phases should remain as they are. [38]

## 3.1.3 ADVANTAGES AND DISADVANTAGES OF WATERFALL MODEL

If the definition of the project is stable, the requirements can be documented precisely. One of the critical things of the model is the predefinition of the technology stack that makes it delicate. Also, there are no ambiguous requirements for the model if the assigned project is short.

| Advantages | Disadvantages |
|---|---|
| The usefulness of the Waterfall SDLC model has its uniqueness in utilising this most quickly and efficiently that makes it widely used SDLC model to serve the purpose for the researchers. | One of the disadvantages which hurts the researchers after the adaption of this model is the readiness of the model. This model is not ready to use for performing practical matters until and unless the last stage of this model is completed. If any work is left to be completed, the model refuses to work according to the purposes it serves with. |
| It is essential to know about the result for the researchers in each phase so that they can be well familiarised with the likely output that it will end up producing. This model enables the researchers to have a view on the output after the completion of each phase. Also, a review of the process of how it has been done- is paved after the phase has been done. | This model is equipped with high uncertainty and possesses a risk to the researcher due to a lack of critical functionality in the model. This model has limited functionality within which it works well and cannot push boundaries to work in smarter ways. |
| The development stage starts after one stage is finished. In this way, the researchers can have all the views before the initiation of another stage. | This model is beneficial for easy and short-term projects. However, this model is not very useful for complex projects and cannot deal with complexities efficiently. |
| This model is suitable for projects that have clear instructions that have fewer functionalities and complexities. Thus, it is better to undertake this model when the project is not so big in nature. | It is not suggested to undertake the model if the project requires numerous functionalities and have complex work abilities required. Hence, small and medium projects can embrace the model, whereas big projects should not embrace this model. |

| In the development cycle of this model, it has the potential to determine the critical points for the model. | On another note, it is difficult to measure the progress while the model has lied in the development stage. |
|---|---|
| Classification and prioritisation of the tasks can take place in the model since the model is segregated into different folds | One of the major loopholes of the model is the inability to identify a problem and rectify the problem accordingly in advance. Hence, the integration is done at the end of the model, which consumes time and effort for the researchers to integrate. |

## 3.2 DATA COLLECTION REQUIREMENTS

Data collection methodology is needed to be formulated since the data collection can differ from different cases. Data can be collected through qualitative or quantitative analysis, or a mixture of both can be done as well. However, this project is based on a product. It is important to have a mixture of both qualitative analyses from the literature review and quantitative analysis from the customers' perception of the product. Hence, data collection methodology will occupy the following methods to collect data from the respondents.

### 3.2.1 INTERVIEW

It is a technique of data collection methodology in which the researchers prepare specific questions to ask the respondents and the respondents are asked to answer the questions. In this method, data collection might take time, but the accuracy of the collected data is authentic, and respondents feel the urge to provide accurate information if they are not being probed. [15]

In our purpose for research, we have conducted interviews with our existing customers with their regular basis queries. And some time with the other customers as well. Based on the interviews, we have built the chatbot training module and trained the chatbot. We have included the discussions in the [APPENDIX A and APPENDIX B]

## 3.2.2 SURVEY

Survey data collection process is considered as the backbone of primary research. It is a medium of data collection which entails the researchers to undertake different types of research interests combined into one so that they can gather data on different bases from the same respondents. So, data collecting a large pool of data may be possible upon embracing this procedure. Therefore, this project can surely be served with the offerings of this data collection procedure and will surely be benefitted too.

Over the project, we have taken multiple approaches to collect data. We have focused on collecting data by the mean of the survey rather than other means. Questions that have been formulated from the survey are analysed in the analysis part of the thesis.

## 3.3 SOFTWARE METHODOLOGY

In order to have completion of this project, numerous types of software methodologies have been studied. There are ample software tools in the market to be used in this regard. However, not all software tools can be susceptible to the needs of this project. After going through numerous software technological tools, pycharm has been considered as IDE with oracles development kit due to the offerings of different benefits. In order to have a fast application as well as a reliable one, the pycharm is one of the leading most widely used software tools to provide support to the researchers or practitioners in that field. Different types of IDE, such as Wing IDE, Eclipse, and pyDev, could have been chosen for the sake of the project. Since acquiring efficiency is one of the top priorities of this project, pycharm has been chosen so that reliability, as well as speed, can be ensured, which are not ensured in other IDEs.

- **MySQL DB**: We have given more priority than other SQL server-based database because of the scalability, performance, reliability and ease of the use. It is the most popular database in the word now
- **Sublime Text 2**: We have used the sublime text editor overriding other popular text editors. There is some reason that we have chosen sublime text over others. Reasons are stableness, customisable, featureful, cross-platform.
- **PhpMyadmin:** is a free and open-source tool for administrative purpose in web development.
- **Cpanel:** Multiple features that can be benefited if cPanel is used in the hosting service, like creating / deleting / forwarding / automatic spam control for email accounts.

Moreover, cloning website, installation of any app in the native language, file restoration, backing up. It is featured options that we cannot get through other service providers.

- **Oracle VM VirtualBox:** It is a free and open-source tool to experiment with different hard things that might mess up with the situation of the hosting computer.
- **Nginx Server:** we have chosen to user Nginx server as this is the best in the market for its advanced features like media streaming, reverse proxying for HTTP protocol. On the other hand, it also gives extra high performance for the dynamic website.

## 3.4 HARDWARE METHODOLOGY

If an eye is put on the methodological part of the hardware, it is evident that this project is not based upon on the core of hardware. Also, the chatbot will not be dependent on any types of hardware too. However, still, as we have used the computer for the development of the chatbot, so we include the hardware part. The specifications of the computer hardware are given as follows:

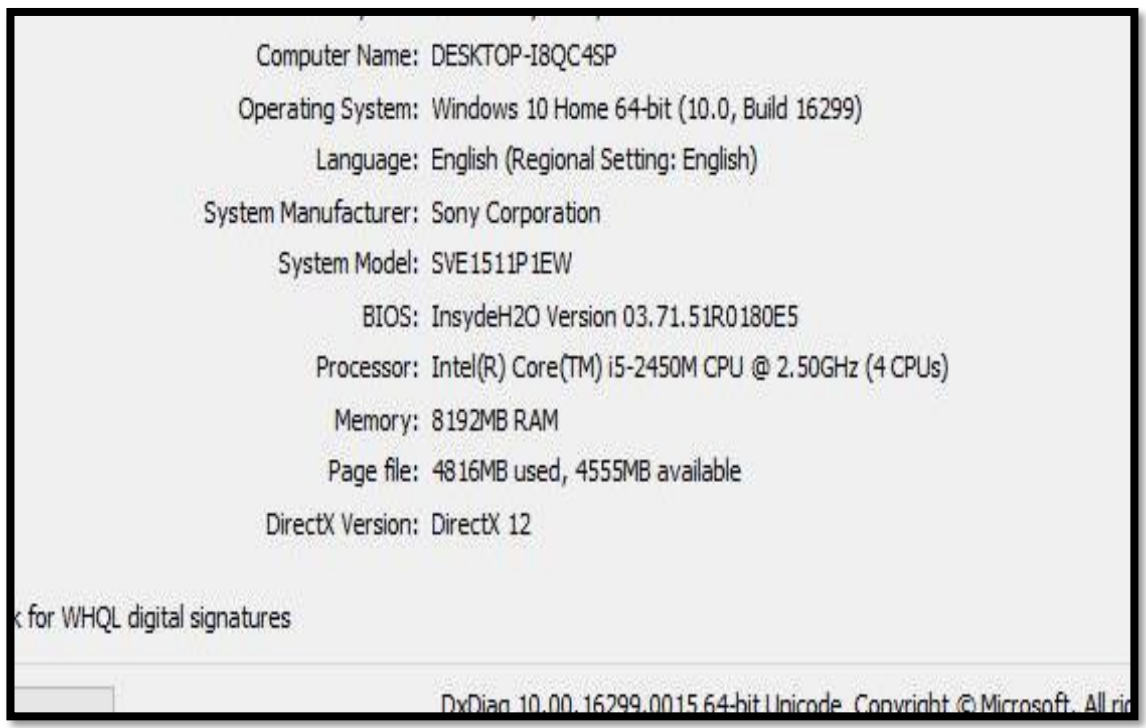

*Figure 11: Configuration of The Computer*

## 3.5 FUNCTIONAL REQUIREMENTS

- Permission to use the application to the unregistered users
- Assisting users to satisfy their queries
- Conversing with the users through text commands
- Understanding the users through natural language
- Staying on a conversational state even if the chatbot might not understand what the users say and should ask for more detailed information

## 3.6 NON-FUNCTIONAL REQUIREMENTS

- Efficient in responding to fast messages
- Should not take more than 5 seconds to reply to the users
- Should be bugs free and reliable to the users
- Scalable database to provide services to numerous numbers of users
- Should be secured due to having sensitive information from the users
- Two-factor authentication system should be introduced in order to provide security measurement intact. We did not implement this feature now, but that could be done in future.
- Should have compliance with the GDPR, which is a form of data protection law
- Promotion of human-computer interaction using natural language as a medium of communication
- Providing accurate and definite responses to the queries of the users
- Taking care of unexpected inputs by providing corrective measures to the users to assist efficiently and effectively

# 4
## CHAPTER

# ANALYSIS

**INFORMATION IN THIS CHAPTER**

- ❖ OVERVIEW
- ❖ ANY BUSINESS COMMUNICATION IN THE LAST 12 MONTHS?
- ❖ ANY ONLINE BASED SERVICES OVER THE 12 LAST MONTH?
- ❖ WHAT IS THE FUTURE OF CHATBOT?
- ❖ ADVANTAGES OF THE CHATBOT SYSTEM?
- ❖ WHY WOULD YOU NOT CONSIDER CHATBOT SYSTEM?
- ❖ CHATBOT VS EMAIL
- ❖ CHATBOT VS PHONE
- ❖ CHATBOT VS APP

# 4.1 OVERVIEW OF ANALYSIS

In our case of data collection, we have adopted two ways, i.e. interviews and surveys. For this data collection methodologies, we have selected a few people who can help us in surveys and interviews. The group of people we have chosen are aged between 19-63, and they are total in 25 in number, and we have done this survey by using Survey Monkey on 05$^{th}$ May 2019. Following are the questions and the response from the participants:

# 4.1 WHAT MEANS HAVE YOU TAKEN FOR BUSINESS COMMUNICATION IN THE LAST 12 MONTHS?

Following bar graph displays the response from the participants:

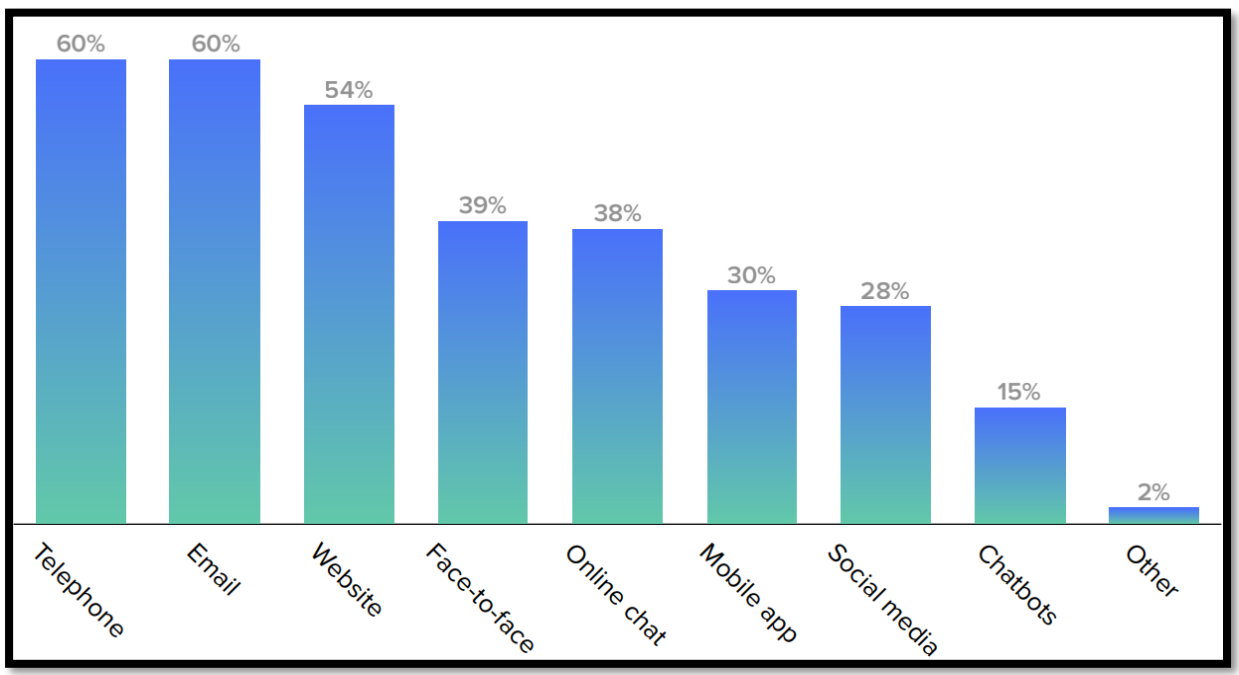

*Figure 12: Different Means of Communication*

The above result from the survey question shows that 15 % of participants have experienced chatbot system to communicate with the business over the given period, i.e. 12 months. However, it can also be seen that there are around one-quarter of the people who have used telephones and email medium for business communication; it's quite a significant amount. We can predict the increasing number of users in this category over the coming years.

Based on the analysis above, we tried to figure out the reasons behind the fact that people are not pondering the chatbot as the medium for business communication.

A chatbot is one of the emerging technologies which is still in its initial stages of development and efficiency to provid better customer experience while answering their queries. To have some more understanding, we advised our participants to take into account of other online-based services, which they frequently use in their daily life, which includes search-engines, various websites and other apps etc.

## 4.2 WHAT WAS YOUR WORST EXPERIENCE WHILE USING ONLINE BASED SERVICES OVER THE LAST MONTH?

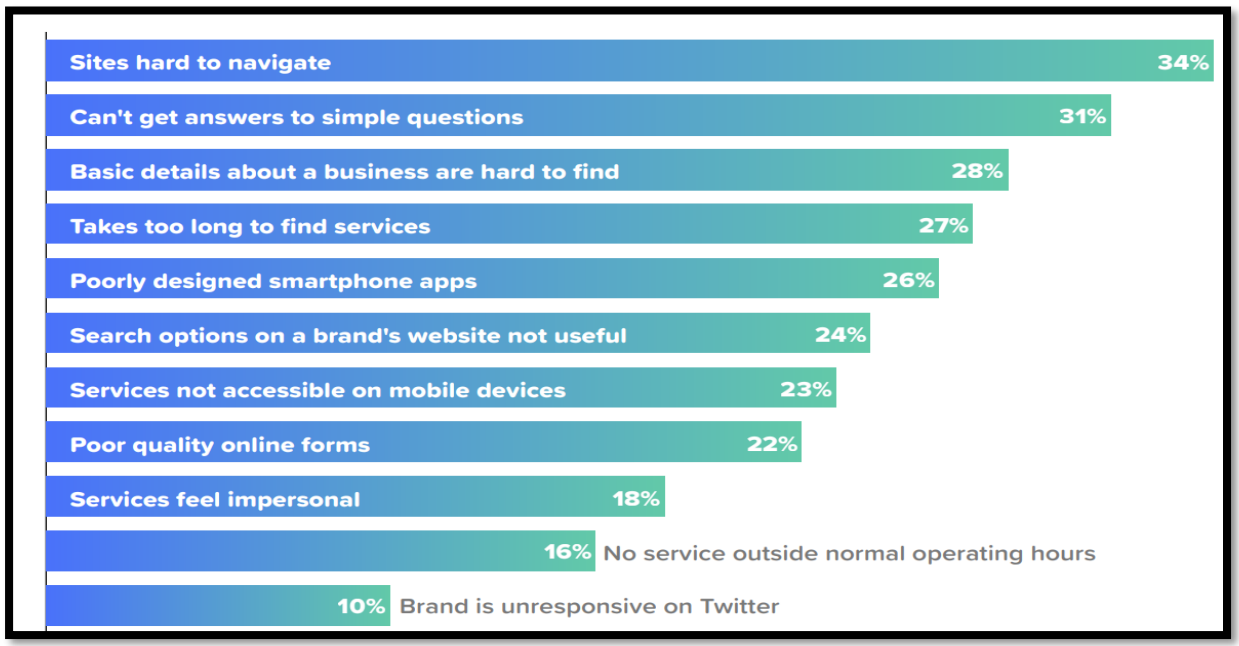

*Figure 13: Online Experience Survey Result*

According to the result shown above on the bar graph, it can be seen that the worst experience of around 34 % of the participant is their struggle to site navigations. Around 31 % considered as not getting the solutions to their answers and around 28 % claiming the necessary business details are harder to find out.

So, it can be concluded from the above result that the people experience in online business communication is quite worse, in the way that, user prefer information from the support team, this means the online system is not efficient to meet the expectations of the users.

Hence, the users are looking for the instant services with quality, when the companies are not able to provide the information as expected in online platform, the users prefer to move to other competitors with the expectations of better services.

So, this need of users triggers to find some efficient solution which is instant and qualitative, and here comes the concept of using machine learning technology in the chatbot.

While continuing the survey with the same participants, we have asked the participants following question to analyse the importance of chatbot system:

## 4.3 WHY DO YOU THINK CHATBOT SYSTEM SHOULD BE USED IN THE FUTURE?

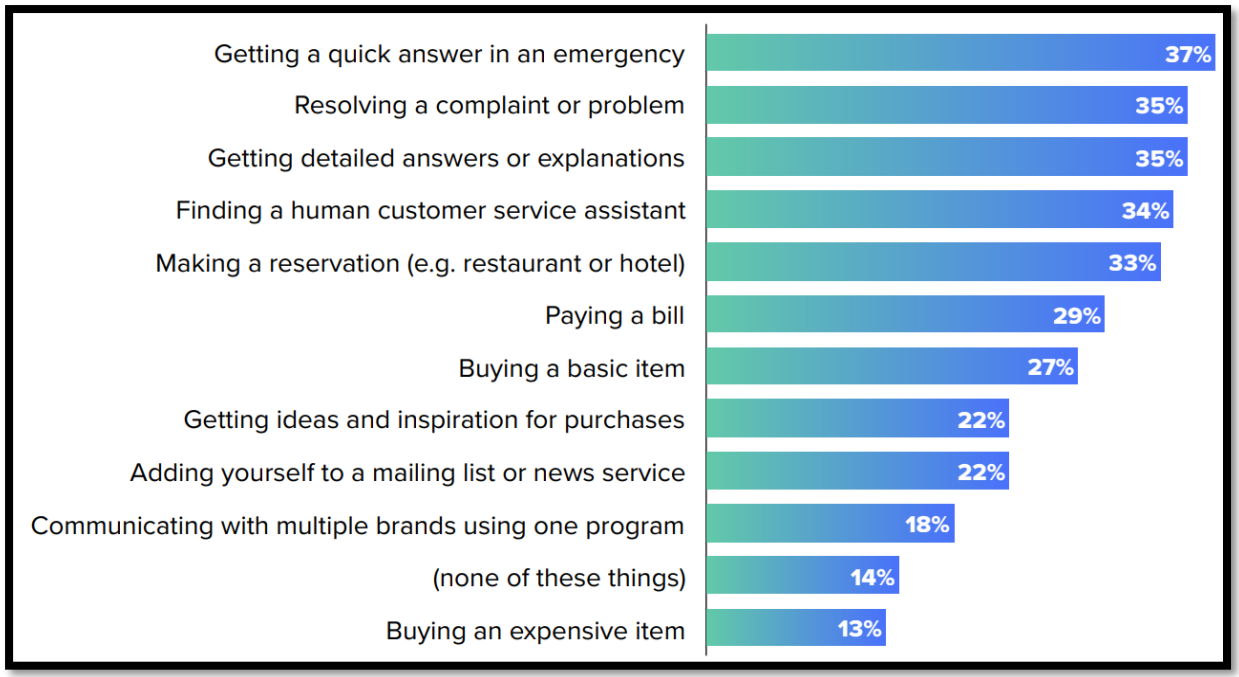

*Figure 14:Results of the survey on the Future of Chatbot*

The bar graph displays the result from the survey mentioned above question. It can be seen that the highest number of participants (i.e. 37 %) prefer chatbot in the future to get an instant solution to their queries. However, 35 % of participants also think that it could be used to resolve issues and complaints, while the same number of participants thinks it could help geta detailed explanation of their question. Interestingly, 34% of participants refer chatbot just to get connected with the human service assistant.

These results show that the users mostly concerned for the chatbot system because of its capability for the instant replay and use it to get some general information. However, for any complex queries or making any purchase, most users prefer to get connected to the human assistant.

While contemplating the above assumptions and results, we have continued survey by asking the following question to know more about chatbot system in user's perspective. Let us find out the chatbot system is working efficiently on the online platforms of your regular use,

## 4.4 WHICH ONE THE FOLLOWING ADVANTAGES WOULD YOU CONSIDER FOR THE CHATBOT SYSTEM?

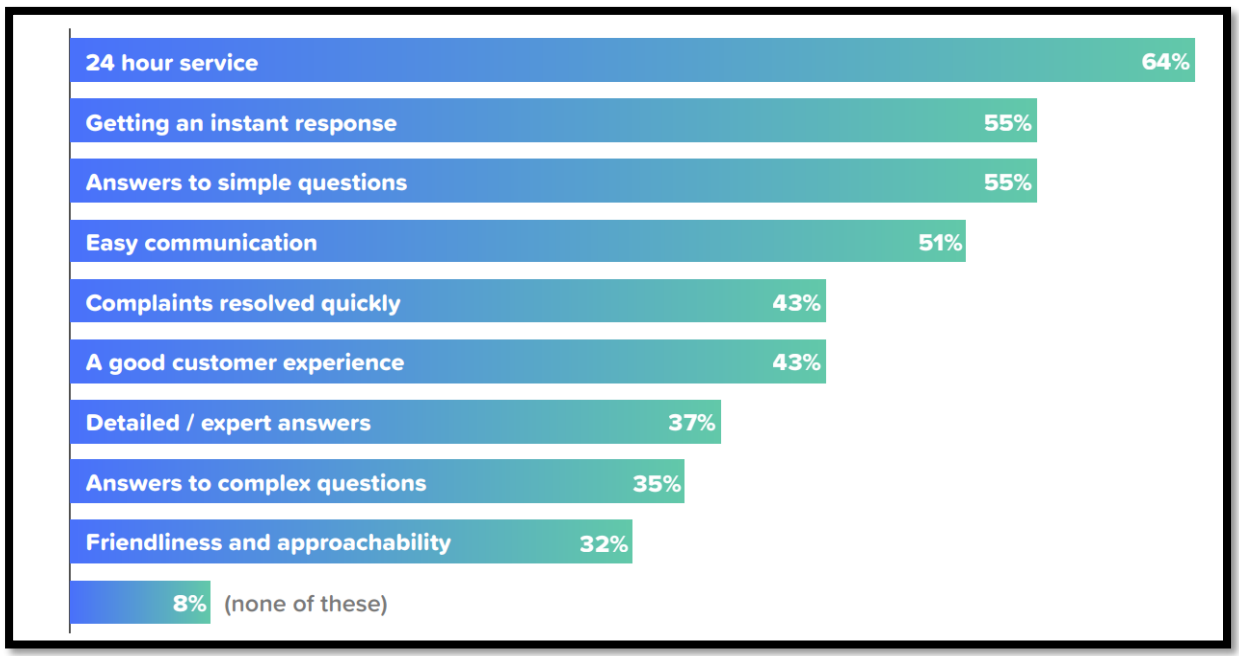

*Figure 15: Advantages of Chatbot System*

The above bar graph displays the result for counting chatbot for online services. The result shows that most of the participants (i.e. 64%) will contemplate chatbot advantage of providing 24-hour availability. In contrast, around 55% of participants will think about the benefit of this technology by instant response generation, and the same participants think that it can be quite useful for giving simple answers to the questions.

We have continued asking more expected experience from the participants by asking the following question:

## 4.5 WHY WOULD YOU NOT CONSIDER CHATBOTSYSTEM?

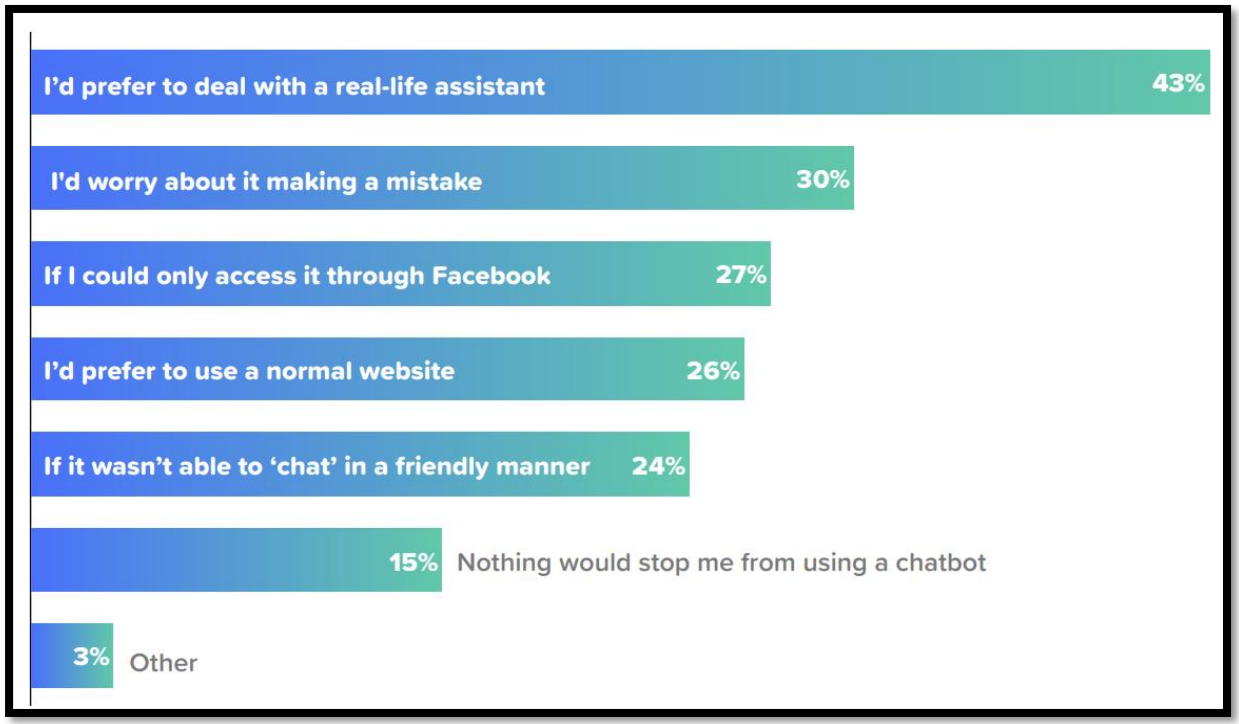

*Figure 16: Results of the "Why you will consider Chatbot?"*

The above bar graph shows the result for not considering chatbot system; it can be seen that the more significant number of participant (i.e. 43 %) prefer to speak to real human than rely on technology for their concern. 30 % of participants are not confident as they think it can cause a mistake. Some say (27 %) to consider chatbot just for social media, i.e. Facebook. One of the impressive figures we found was 15 % of the participant said that there is no reason for not taking the chatbot system into account, and this number was equally from all class of ages.

The result shows that people are confused in various ways, the primary reason they are not fully confident in the technology. In other words, there is no awareness that exists in people which could make them understand the potential benefits of this technology. The main point for any business to consider the chatbot system is to ensure the instant replies and for any other complicated situation, a chatbot could refer to the human assistant.

We have asked the participant comparing chatbot vs email, Chatbot vs app and chatbot vs phone. The question was:

## 4.6 WHAT ARE THE ADVANTAGES DO YOU FACE WHILE DOING BUSINESS COMMUNICATION?

### 4.6.1 CHATBOT VS EMAIL

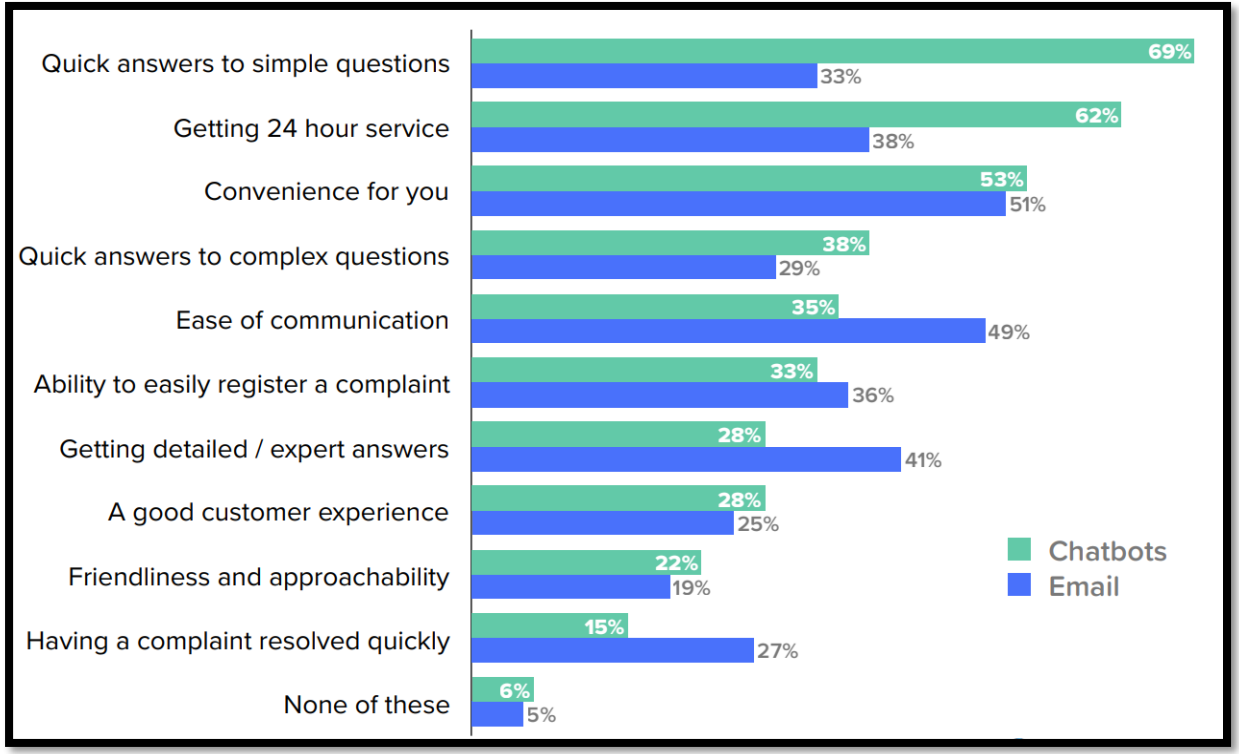

*Figure 17: Chatbot vs Email*

## 4.6.2 CHATBOT VS PHONE

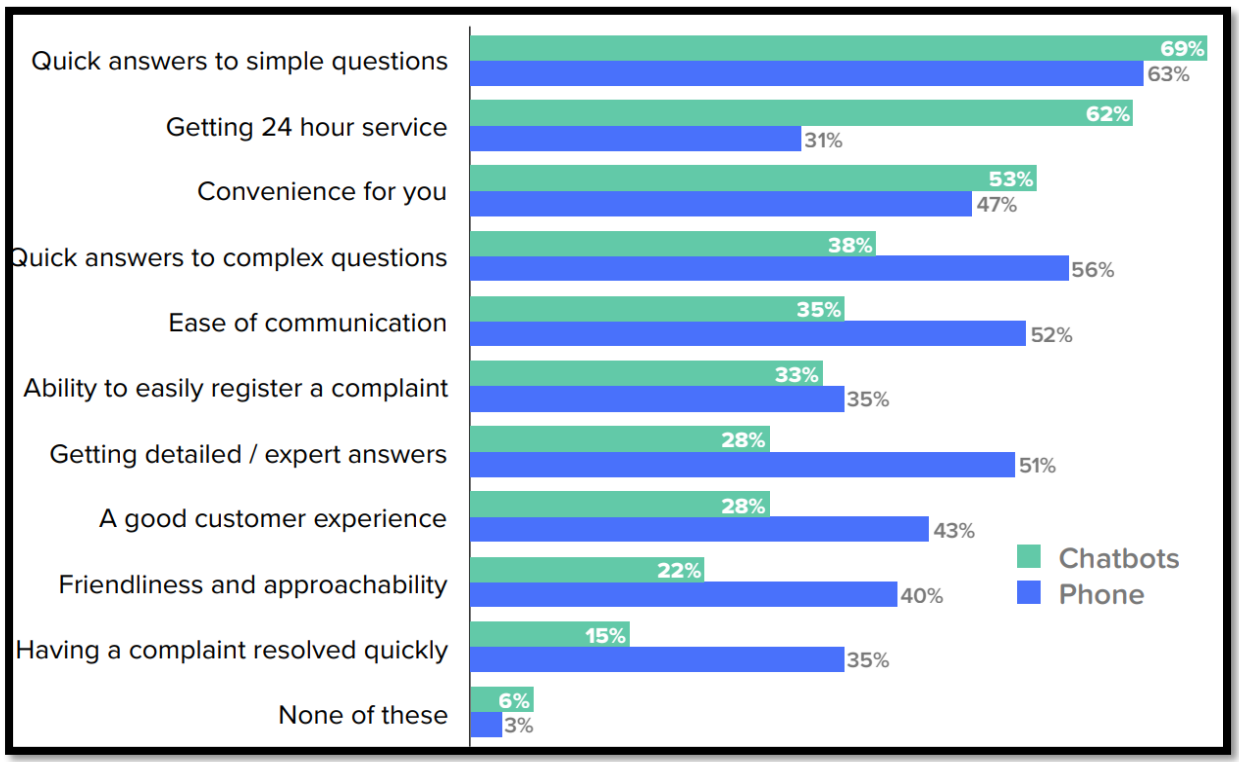

*Figure 18: Chatbot vs Phone*

## 4.6.3 CHATBOT VS APP

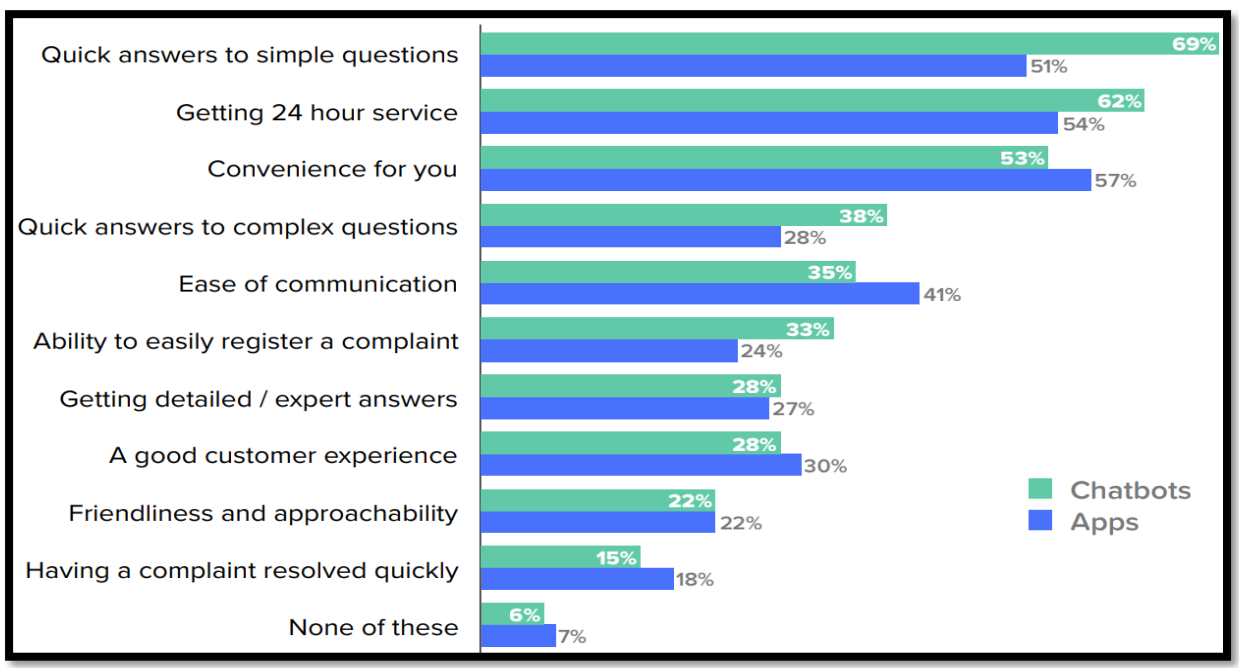

*Figure 19:Chatbot vs Apps*

While considering the business communication, the chatbot is still underway of replacing the use of phones and emails. It can be seen from the above results, that while comparing the chatbot with the apps, participant preferred chatbot over the app in the five various ways which includes instant replies, 24 hours access, detailed explanation and suggestion the solution to complex questions. Furthermore, if we compare it to the email section, it can be seen that the participant preferred chatbot over the email in an instant replying and 24-hour service. If we look at results by comparing with the phone, it is worth to note that participant preferred the phone for an instant solution over the chatbot. Email and phones are the primary media to resolve complex issues and to register complaints.

It can be concluded from the survey that chatbot is still at the stage of development and improvement for processing complex queries; however, it is still considered one of the best media for business communication when it comes to instant replies and 24-hour service.

# 5

**CHAPTER**

# ARTEFACT DESIGN

## INFORMATION IN THIS CHAPTER

- ❖ OVERVIEW
- ❖ ANY BUSINESS COMMUNICATION IN THE LAST 12 MONTHS?
- ❖ ANY ONLINE BASED SERVICES OVER THE 12 LAST MONTH?
- ❖ WHAT IS THE FUTURE OF CHATBOT?
- ❖ ADVANTAGES OF THE CHATBOT SYSTEM?
- ❖ WHY WOULD YOU NOT CONSIDER CHATBOT SYSTEM?
- ❖ CHATBOT VS EMAIL
- ❖ CHATBOT VS PHONE
- ❖ CHATBOT VS APP

# 5.1 OVERVIEW OF DESIGN

In this chapter, we had taken deliberation of the system design, its process of developing. We also include the diagrams that give the overview of the design of the chatterbot. Gantt chart is also explained in this section to illustrate the time frame we have used throughout the development process.

# 5.2 BASIC CHATBOT DIAGRAM

The chatbot is based on the customer service, so the datasets such as question-answer as well as the question-to-question are based on the one to one conversation for customer service. However, there are some models which can have different applicability to the different domains. The chatbot is supposed to be general since it can have applicability in different domains for different datasets such as question-answer datasets and question-to-question datasets. Different methods along with different models have been explained and from the discussion took place in prior sections, the chatbot has been devised by using machine learning which will be discussed vividly in later sections of this report.

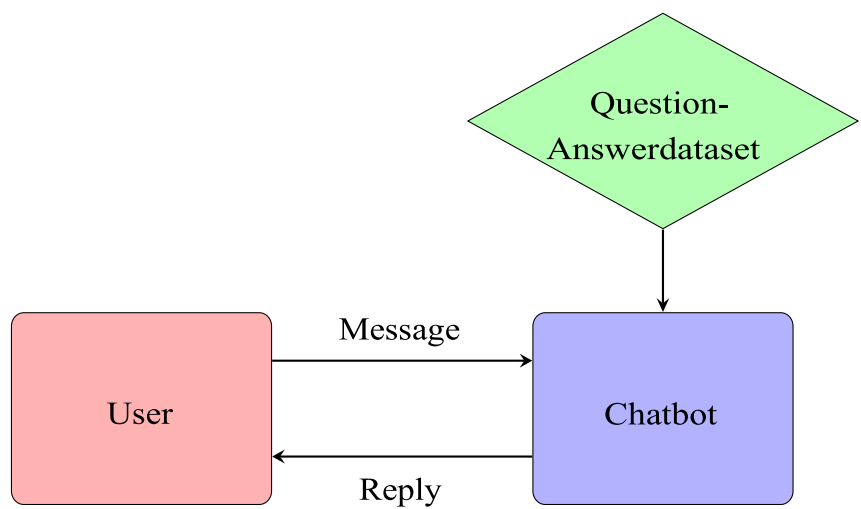

*Figure 20: Basic Diagram Of Chatbot*

This diagram provides a complete general view of the overall functionalities of the machine learning chatbot model.

## 5.3 FLOW DIAGRAM DESIGN

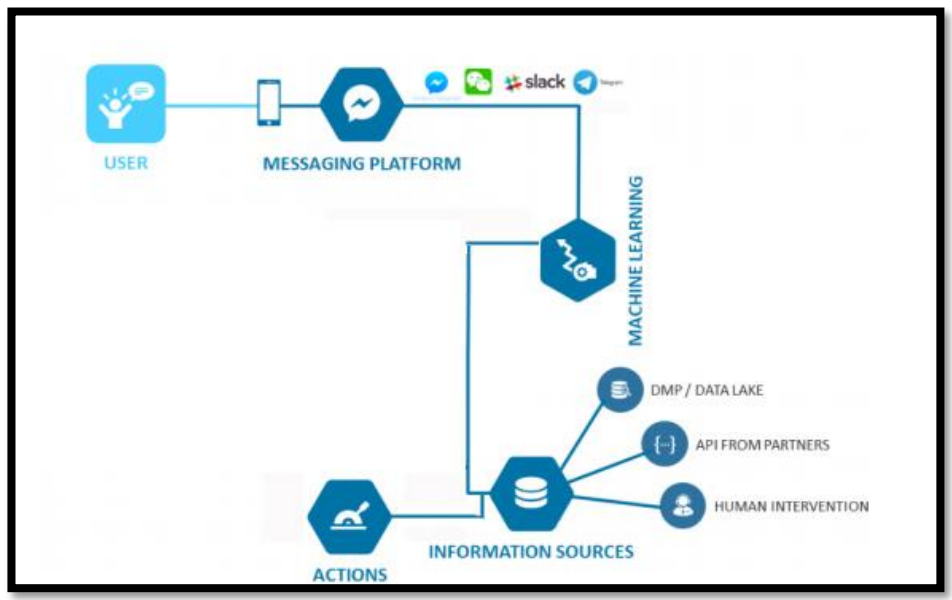

*Figure 21: Flow diagram of your chatbot*

Flow Diagram is a visual representation of the sequence of steps and decisions needed to perform a process each step in the sequence is noted with the diagram shapes. Steps are linked by connecting lines, so that anyone to view the Diagram could logically follow the process from beginning to end of the flow. It is a powerful tool to design and construct the necessary steps in the process very effectively notice the diagram with different shapes.[17]

Here we can see that the user directly connects with the chatting platform, either with a third-party platform or made the platform like Facebook messenger/slack or another mean. That platform connects with the machine learning algorithm. Machine learning codes and algorithm connects with the API, database, and it acts accordingly.

## 5.4 ACTIVITY DIAGRAMS

The activity diagram is responsible for depicting the flow of information. In addition, it can also provide an overview of all the processes which contribute to the chatbot. The diagrams provide a visual representation of the behaviour of the users when the users try to communicate with the chatbot. The diagrams also possess the distinctive workability of the chatbot. Such as when a user asks for any query of the taxi, suggests the logic of chatbot meanwhile. After the evaluation of the user's input by a machine learning algorithm, the server comes into the scenario and works for the POST endpoints. Also, the server approves the requests which have already got back from machine learning algorithm. Afterwards, the server should engage in determining the intent of the users.

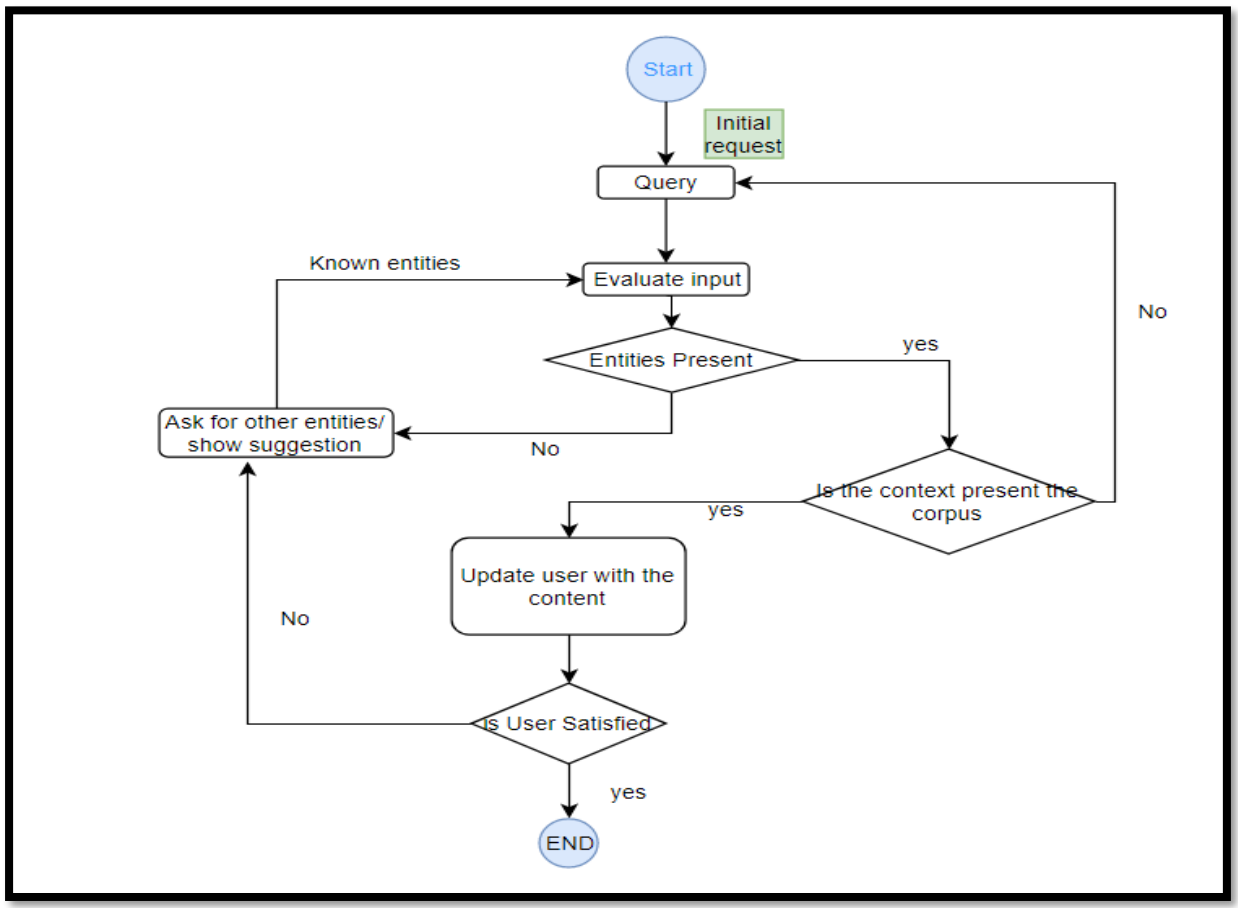

*Figure 22: Activity diagram*

In the user starts the conversation with the Chatbot with their query. If the question is relating with the conversation that is already in the machine learning algorithm, then it will represent with the response from the server, if not, then it will give a suggestion or ask for new entities.

## 5.5 SEQUENCE DIAGRAMS

The diagram is known as a sequence diagram. It represents the procedure of a user when he/she asks for any information by using the chatbot. This is a fundamental and primary process involved in the beginning. In the form of request, the server initiates the arrival from the chatbot machine learning algorithm engine. Later, the server involves in determining the actions which are needed in order to analyse the asking it has already received. Afterwards, the server will undertake necessary steps if the action can have any match with the customer service. after receiving the data which has already been returned by API, the data encoding begins and that is the response which is delivered back to the machine learning algorithm. Then the user can have a view through the chatbot.

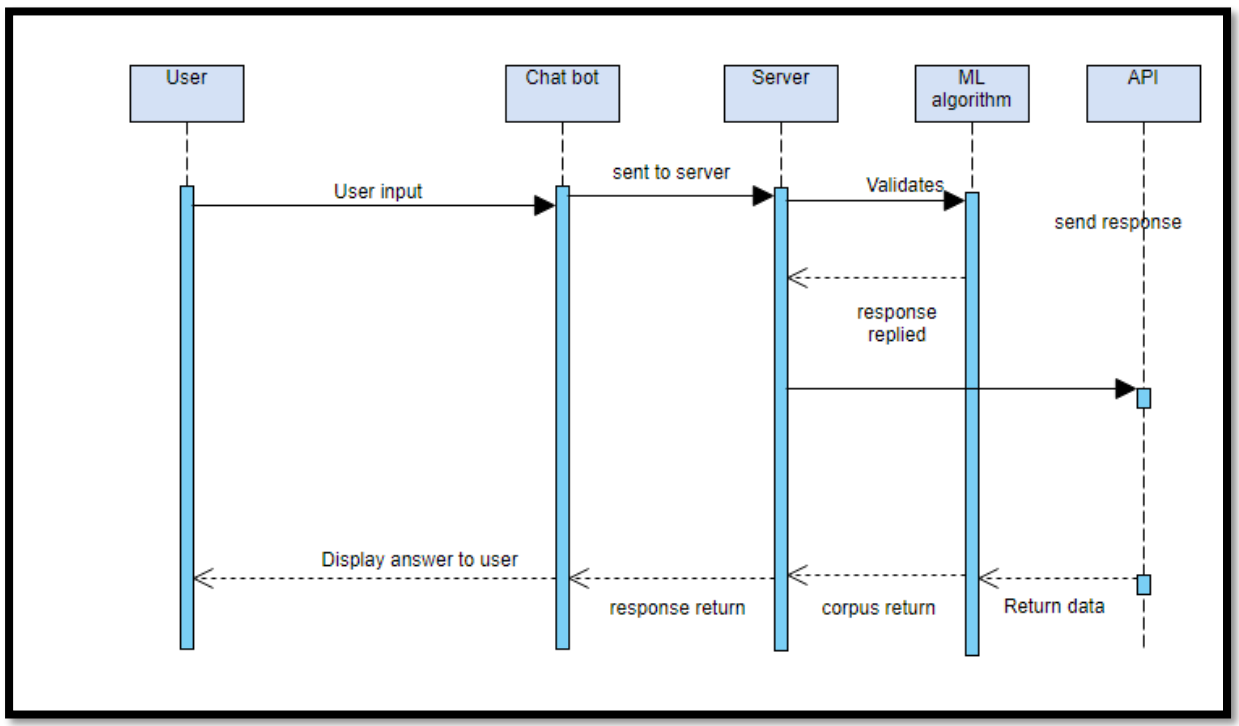

*Figure 23:Sequence diagram*

## 5.6 PROCESS FLOW DIAGRAM

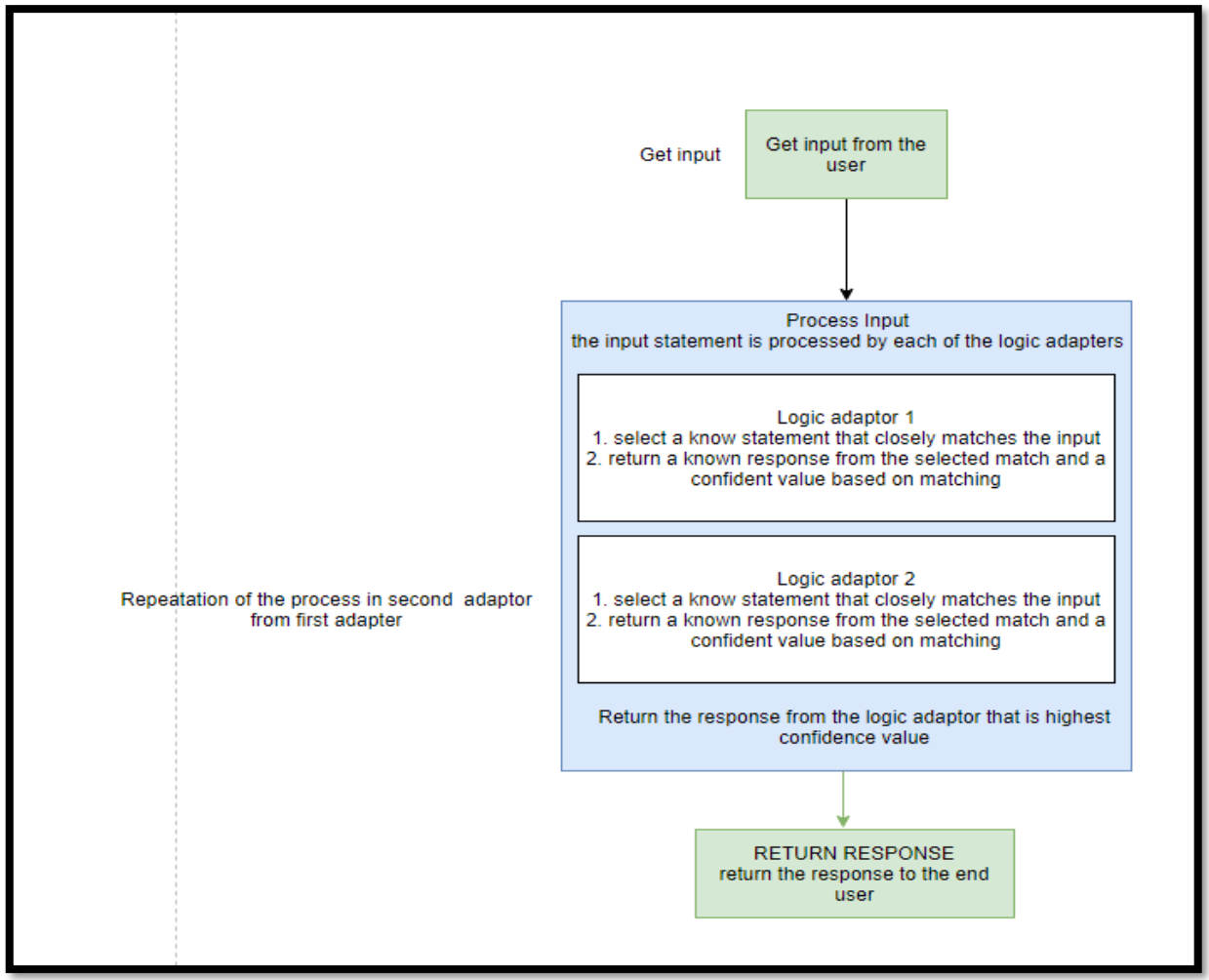

*Figure 24:Process Flow Diagram*

An untrained instance that has no prior knowledge about how to make conversation learns the conversation from the training module of the chatterbot library. As much as we train, the chatterbot learns how to respond with a faster and more reliable match. Accuracy of each response gets better and better. The above diagram shows the process flow that the instance uses in every conversation.

## 5.7 ARCHITECTURAL DESIGN

The figure portrays an architectural design of the proposed chatbot.

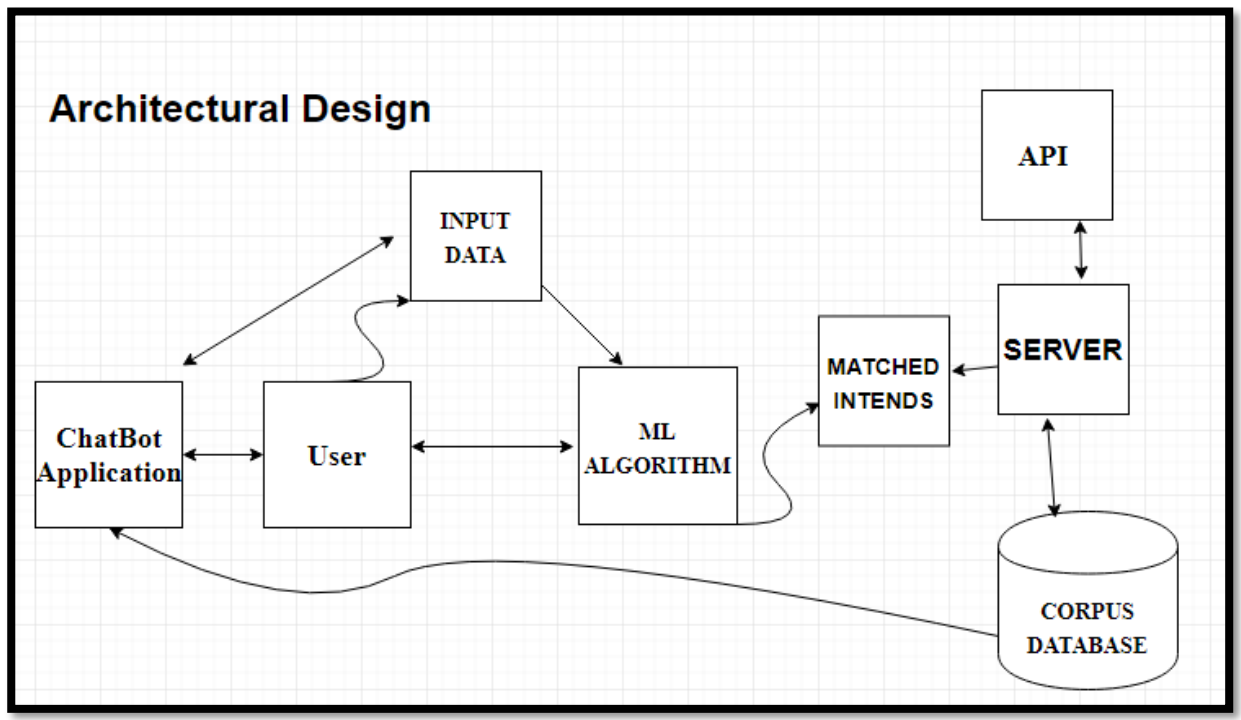

*Figure 25: Architectural design*

Through the web client, the users will be communicating with the chatbot. The medium of communication between the user and the chatbot will be natural language. CSS, HTML5 and JavaScript have been used to formulate the client side of this application. Besides, this diagram allows us how to utilise in executing the web client, chatbot. The server is entitled to receive the requested data through HTTPS POST from a machine learning algorithm. The route can easily be stated as the endpoint since machine learning algorithm can post real-time payload. In addition to that responsibility, the server also receives data when a user has already established intent. Therefore, the server can be able to receive different data which has been posted from different service at the time of occurring any event apart from this [4].

This diagram shows the chatbot to be familiarised with the intents as well as entities by providing different training and is dedicated to utilising the machine learning algorithm engine. The chatbot will have the structure based on the intent map to route the utterances for the users to the consolidation of different words.

The figure depicts the information flow, which works like an intent by a user. It is responsible for checking whether the intent was posted or not since it has been trained to do so. Later, the data should be posted on the server. A response is then originated, which is custom in nature.

## 5.8 PROJECT MANAGEMENT

Project management is essential in term of any project that is conducted in a supervised environment. We have used a Gantt chart to show the timeline that has been utilized in this project development.

### 5.8.1 GANTT CHART

A Gantt chart has been formulated in order to provide a portrait on the timeline of the project. According to the following Gantt chart given below, the project is expected to have completed on time. Similarly, from the chart, the breaking down of different tasks, the order which should be completed before and after, and the allocation of each task to be completed can be known.

| Tasks | Duration | Start Date | End Date |
|---|---|---|---|
| Job InterView | 1 | 25/03/2019 | 25/03/2019 |
| Job Duration | 119 | 01/04/2019 | 22/08/2019 |
| Project Proposal | 1 | 24/03/2019 | 24/03/2019 |
| Placement Acceptance by university | 1 | 02/04/2019 | 02/04/2019 |
| Discuss the Job role with Work Supervisor | 1 | 03/04/2019 | 03/04/2019 |
| Design of poster, leaflet / writing content etc | 112 | 10/04/2019 | 22/08/2019 |
| Define and outline the problem | 21 | 17/04/2019 | 08/05/2019 |
| survey | 28 | 13/05/2019 | 05/06/2019 |
| Interview | 35 | 10/06/2019 | 19/06/2019 |
| Requirement Analysis | 42 | 24/06/2019 | 01/07/2019 |
| Litreture Review | 49 | 01/07/2019 | 29/07/2019 |
| Feasibility Study | 56 | 29/07/2019 | 05/08/2019 |
| Anaysis, requirements and specifications | 63 | 05/08/2019 | 02/09/2019 |
| chatbot design and development | 70 | 03/07/2019 | 23/10/2019 |
| Implementation | 77 | 19/08/2019 | 08/10/2019 |
| Testing | 84 | 09/10/2019 | 23/10/2019 |
| Evalutation | 91 | 21/10/2019 | 28/10/2019 |
| Finalizing of Whole project | 98 | 28/10/2019 | 04/11/2019 |
| Report Writing | 105 | 22/08/2019 | 15/11/2019 |

*Figure 26: Gantt Chart*

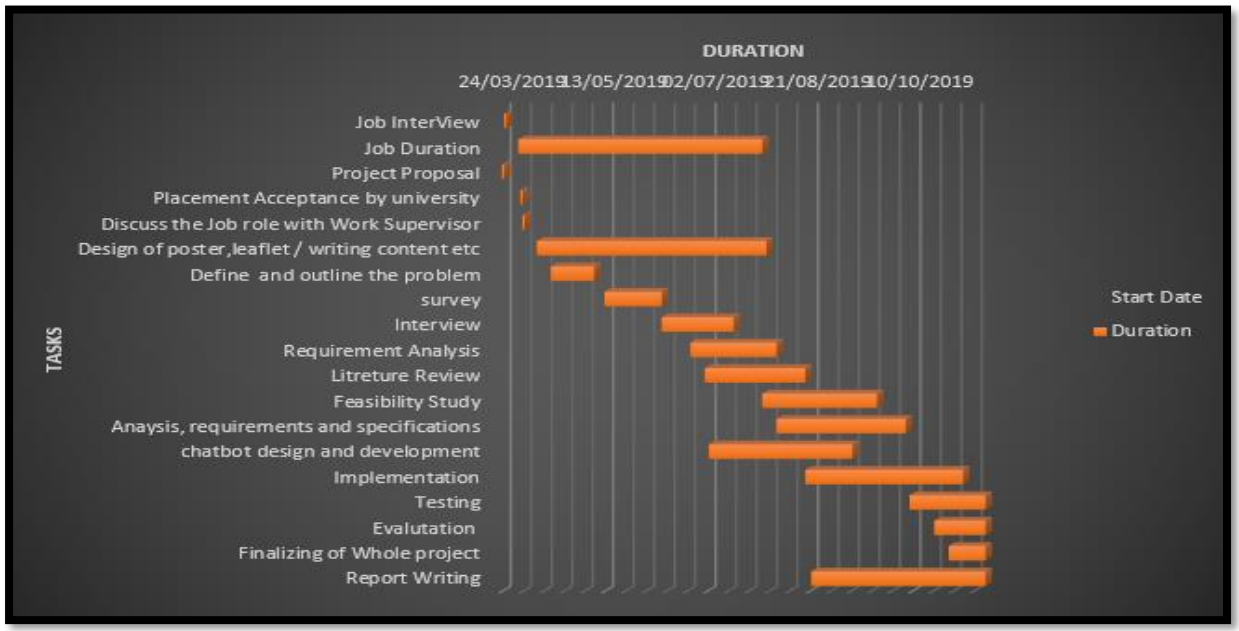

*Figure 27: Graph Gantt Chart*

The graph Gantt chart illustrated the time duration that has been carried out during the whole project period from the beginning of the placement and the end of the report writing. It also clearly shows the duration taken for a particular task and the start date to the end date. It also shows which task took more time than others. We can see that chatbot design and development along with the implementation took more time than any other tasks

## 5.8.2 RISK MANAGEMENT

To eliminate the risks and the end consequences of any project, we need to identify and manage the risk concerned with the project. It is an essential step for any project's overall progression impact. Analysing and knowing the risks repercussions at the very beginning of the project minimises the occurrence of the risk. It gives us time to take actions accordingly in the actual risk occurrence. For our project, we take back up a snapshot of the code very often to prevent complete loss of the hard-working codes and the algorithms. Use of google drive made the backup process of the report and the important document accessible. So, we are prepared for the unplanned day.

# 6 IMPLEMENTATION AND TESTING
CHAPTER

## INFORMATION IN THIS CHAPTER

- ❖ OVERVIEW OF IMPLEMENTATION AND TESTING
- ❖ INSTALLATION
- ❖ LIBRARIES AND MODULES
- ❖ PREPARING FOR THE DEPENDENCIES FOR THE CHATBOT
- ❖ TRAINING THE CHATBOT
- ❖ INTEGRATING THE CHATBOT IN THE WEBSITE
- ❖ WEB INTERFACE
- ❖ TESTING
- ❖ ERROR HANDLING

# 6.1 OVERVIEW OF IMPLEMENTATION AND TESTING

Software environment and the context should be considered in order to choose the programming language. In this case, Python has been chosen as a primary programming language so that it can be executed in the chatbot. It is a choice of balance between the future software developer (Who will work after me) and the execution of learning catalogue of the given machine. Whilst finding chatbot for the implementation phase, we had tested a couple of other chatbots, but none worked well in our environment well. [Appendix F, Appendix G and Appendix H]

# 6.2 INSTALLATION

For the chatterbot to work, we need to install some of the prerequisite software in the system. We need to install PyPI, pip, chatterbot some other dependencies.

## 6.2.1 PyPI INSTALLATION

```
rizuan611@ubuntu:~$ sudo apt-get install pypi
[sudo] password for rizuan611:
Reading package lists... Done
Building dependency tree
Reading state information... Done
```

## 6.2.2 PIP INSTALLATION

```
rizuan611@ubuntu:~$ sudo apt-get install pip
Reading package lists... Done
Building dependency tree
Reading state information... Done
E: Unable to locate package pip
```

## 6.2.3 CHATTERBOT INSTALLATION

```
Sudo pip3 install chatterbot
```

```
Requirement already satisfied (use --upgrade to upgrade): nltk<4.0,>=3.2 in /usr
/local/lib/python3.5/dist-packages (from chatterbot)
Requirement already satisfied (use --upgrade to upgrade): chatterbot-corpus<1.1,
>=1.0 in /usr/local/lib/python3.5/dist-packages (from chatterbot)
Requirement already satisfied (use --upgrade to upgrade): sqlalchemy<1.2,>=1.1 i
n /usr/local/lib/python3.5/dist-packages (from chatterbot)
Requirement already satisfied (use --upgrade to upgrade): future in /usr/local/l
ib/python3.5/dist-packages (from python-twitter<4.0,>=3.0->chatterbot)
Requirement already satisfied (use --upgrade to upgrade): requests in /usr/lib/p
ython3/dist-packages (from python-twitter<4.0,>=3.0->chatterbot)
Requirement already satisfied (use --upgrade to upgrade): requests-oauthlib in /
usr/local/lib/python3.5/dist-packages (from python-twitter<4.0,>=3.0->chatterbot
)
Requirement already satisfied (use --upgrade to upgrade): six>=1.5 in /usr/lib/p
ython3/dist-packages (from python-dateutil<2.7,>=2.6->chatterbot)
Requirement already satisfied (use --upgrade to upgrade): ruamel.yaml<=0.15 in /
usr/local/lib/python3.5/dist-packages (from chatterbot-corpus<1.1,>=1.0->chatter
bot)
Requirement already satisfied (use --upgrade to upgrade): oauthlib>=0.6.2 in /us
r/lib/python3/dist-packages (from requests-oauthlib->python-twitter<4.0,>=3.0->c
hatterbot)
You are using pip version 8.1.1, however version 9.0.1 is available.
```

We have installed the chatterbot on our computer.

## 6.3 LIBRARIES AND MODULES

In the term of libraries that we used in building a chatbot, we have considered some libraries that helped us create an intelligent chatbot. The first and essential library we used was the chatterbot library that helps build the autonomous response chatting system for any purpose. Its response based on the user's input. It uses machine learning algorithms to different output kind of response. Another important use is Pandas that gives the developer to keep the data organised and efficiently. This library is used to manipulate and retrieve data from a considerable amount of information stored. Finally, a simple but important module that has been used in the project was random. It is used to get a random number of outputs from a range that is in the same argument.

## 6.4 PREPARING FOR THE DEPENDENCIES FOR THE CHATBOT

After creating a Python virtual environment, we have to install the chatterbot library, so the following commands help us to install.

```
Pip install chatterbot

Pip install chatterbot_corpus
```

It is better to upgrade every time we install anything. So, the command is

```
pip install--upgradechatterbot_corpus

pip install--upgradechatterbot
```

## 6.5 IMPORTING CLASSES

For the purpose of the chatterbot, we will need to import two classes for now, so the command is as follows.

```
From chatterbot import ChatBot

From chatterbot.trainers import ListTrainer
```

## 6.6 TRAINING THE CHATBOT

The ChatterBot has the inclusion of different types of tools so that the whole process of training can be simplified. The dialogue process is involved in different example to put into the database of the chatbots. Known responses are presented, which is created based on the data structure. The entries are created upon providing a data set by the trainer of chatbots. Afterwards, the responses can be represented in the correct manner.

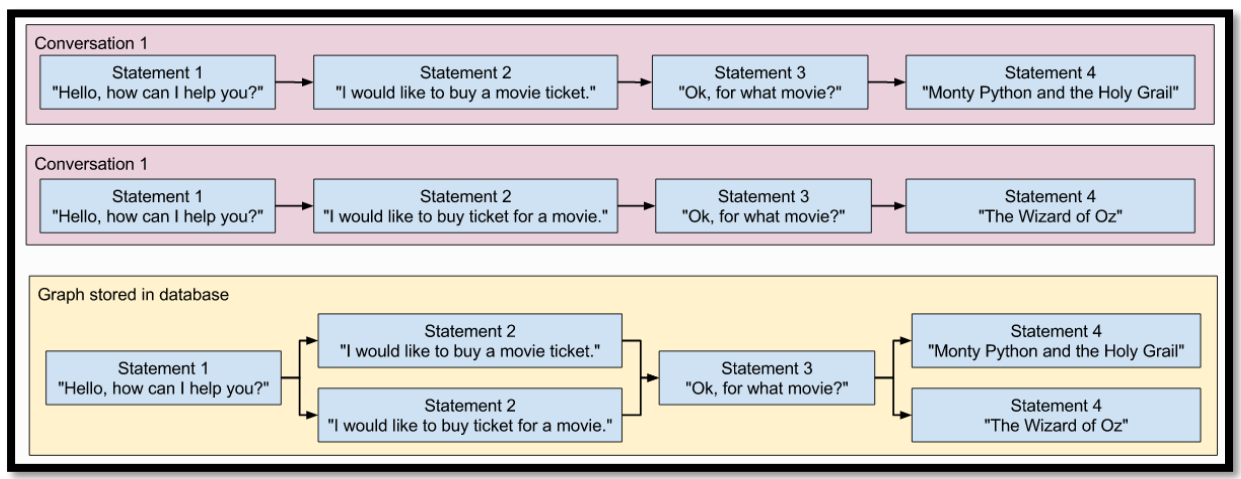

*Figure 28: Training phase in the Graph*

There are a few training classes which are already built-in with the ChatterBot. The utilities can range differently, such as updating the knowledge of the chatbot's database to the permission to allow for having the bot trained. It should be done based on corpus training data that should be pre-loaded by nature. In addition to that, one can devise his/her intended training class. This can be recommended if one intends to train the bots with those data that have been kept in storage and has not been supported by the pre-built classes. The list is given in following:

## 6.6.1 TRAINING CLASSES

A built-in training class accompanies Chatterbot. We can create our training system in the chatbot to train the instance.

Also, it is possible to come up with a new one upon necessity. It is vital to call train () which has been begun with the chatbot to train the chatbot.

At the beginning of the development phase, the instance is not intelligent. It has no clue of the human conversation. The chatterbot has some tools that can process the training phase of the chatterbot. We can train the chatterbot in two main ways

- Training over the list data.
- Training with corpus data.

For learning how the chatbot works we have tested on both the system, but for implementation purpose, we have used the corpus data system to train the chatbot. When the trainer provides training to the chatbot, it achieves the necessary knowledge to respond with the statement inputs and response as represented in the training phase.

## 6.6.2 TRAINING VIA LIST DATA

chatterbot.trainers.ListTrainer(chatbot, **kwargs)[source]

It provides the chatbot with permission to have training upon utilising different types of strings in which a conversation is represented in the list.

Concerning the process of providing training, it is essential to pass a whole list which consists of different statements. The organisation of the statements is done based on a particular conversation.

For example, if one intends to run bot by the saying, "Hello", the chatterbot's respond on the statements will result in, "Hi there!", or "Greetings!".

```
                          chatbot.py

chatbot = ChatBot('Training Example')

                           train.py

from chatbot import chatbot
from chatterbot.trainers import ListTrainer

trainer = ListTrainer(chatbot)

trainer.train([
    "Hi there!",
    "Hello",
])

trainer.train([
    "Greetings!",
    "Hello",
])
```

*Figure 29: Training via list data*

The long list of the training conversations can be provided which can locate every item as a response to the list of all the predecessors.

```
trainer.train([
    "How are you?",
    "I am good.",
    "That is good to hear.",
    "Thank you",
    "You are welcome.",
])
```

*Figure 30: Training via Long list data*

## 6.6.3 TRAINING WITH CORPUS DATA

chatterbot.trainers.ChatterBot CorpusTrainer(chatbot, **kwargs)[source]

This can permit the chatbot to have trained by utilising different data from the dialogue corpus, ChatterBot. ChatterBot relates to the corpus data along with the utility module, which can ensure smooth training to the bot. In order to do this, it is imperative to indicate the corpus module, which is intended to utilise.

```
chatbot.py

chatbot = ChatBot('Training Example')

train.py

from chatbot import chatbot
from chatterbot.trainers import ChatterBotCorpusTrainer

trainer = ChatterBotCorpusTrainer(chatbot)

trainer.train(
    "chatterbot.corpus.english"
)
```

*Figure 31: Training via Corpus*

As we have trained the chatterbot in just a simple chatbot. We feed the data into the system in a .yml file. Sample of the greeting feeding is as below

```
1   categories:
2   - conversations
3   conversations:
4   - - Good morning, how are you?
5     - I am doing well, how about you?
6     - I'm also good.
7     - That's good to hear.
8     - Yes it is.
9   - - Hello
10    - Hi
11    - How are you doing?
12    - I am doing well.
13    - That is good to hear
14    - Yes it is.
15    - Can I help you with anything?
16    - Yes, I have a question.
17    - What is your question?
18    - Could I borrow a cup of sugar?
19    - I'm sorry, but I don't have any.
20    - Thank you anyway
21    - No problem
22  - - How are you doing?
23    - I am doing well, how about you?
24    - I am also good.
25    - That's good.
26  - - Have you heard the news?
27    - What good news?
28  - - What is your favorite book?
29    - I can't read.
30    - So what's your favorite color?
31    - Blue
32  - - Who are you?
33    - Who? Who is but a form following the function of what
34    - What are you then?
35    - A man in a mask.
36    - I can see that.
```

*Figure 32: Corpus data*

If we test the chatbot, we can see that it is responding accordingly.

```
List Trainer: [####################] 100%
You: hi
Bot:  hello

You: how are you
Bot:  I'm fine thanks for asking

You: awesome
Bot:  Yes, how's yours?

You: your what?
Bot:  It was great.

You: What was great
Bot:  Awesome

You: Are you crazy?
Bot:  I'm fine thanks for asking

You: Did you understand this?
Bot:  No, which one?

You:
```

*Figure 33: After Corpus training*

We have built our own corpus to train the chatbot to make the conversation more humanly

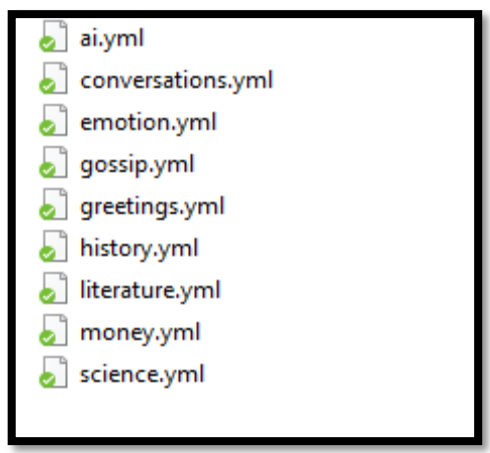

*Figure 34: Different Corpus*

These are the other files that we have used in the chatterbot.

## 6.7 INTEGRATING THE CHATBOT IN THE WEBSITE

After building up the chatbot, we need to deploy the chatterbot into the E20Minicabs website. There exists a variety of web-based framework that can be used for generating projects that are Python-based. The best example of these projects can be considered Django and Flask. The designing of the chatterbot is followed with quite user-friendly steps in terms of its compatibility with the allocated platform. The initial requirement to deploy/run chatterbot on the web-based application is its capability to receive and send incoming data. It can be done in several ways, like with HTTPs, including web-sockets etc. To apply chatterbot in our web application, we have used Django.

We need to install Django and the chatterbot so that the python program can work in webserver

```
pip install Django chatterbot
```

we also need to migrate the database, so the command is

```
python manage.py migrate django_chatterbot
```

## 6.8 WEB INTERFACE

In the following picture, we can see that the web integration is done. The chatterbot is in green colour in the right bottom of the corner of the page

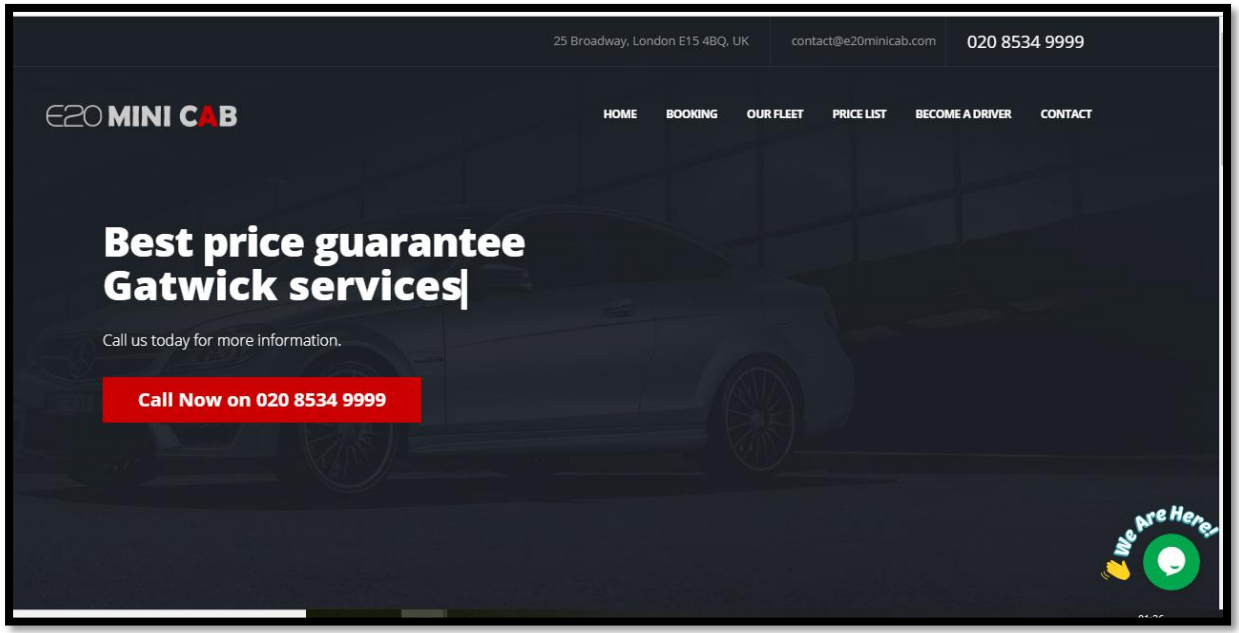

Figure 35: Web Interface

We have designed the chatterbot to popup if the user is spending more than 15 seconds on the website.

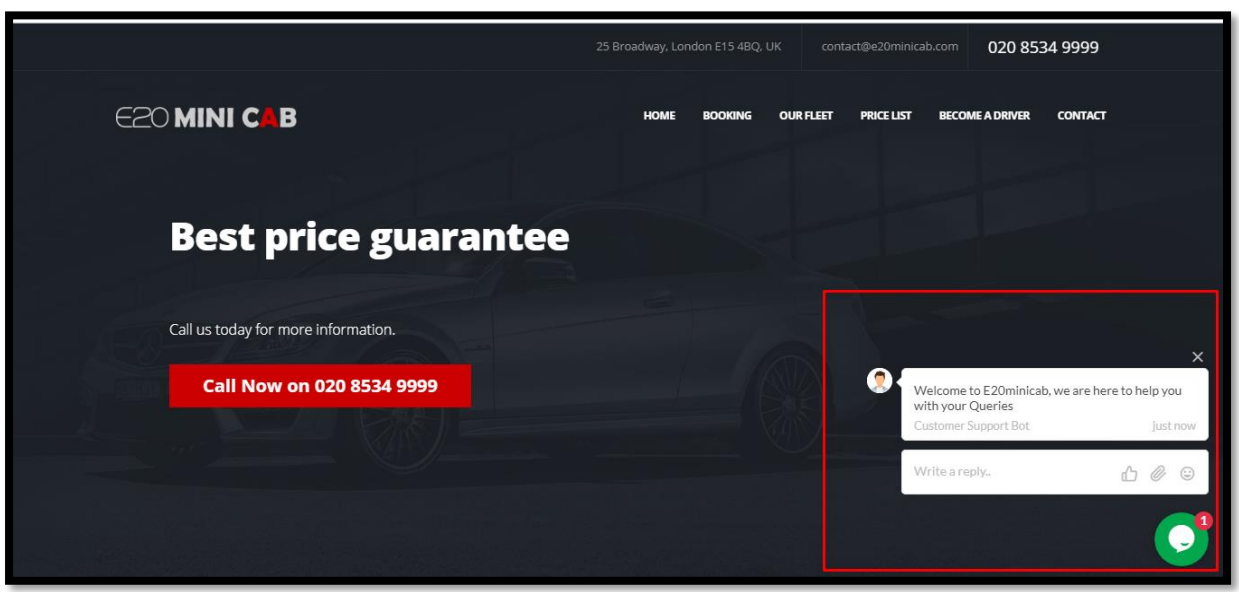

Figure 36: Chatbot Appears

Following screenshot shows the chatbot popping up automatically once web application is launched, and it asks the user for any assistance required and it carries on with chatting:

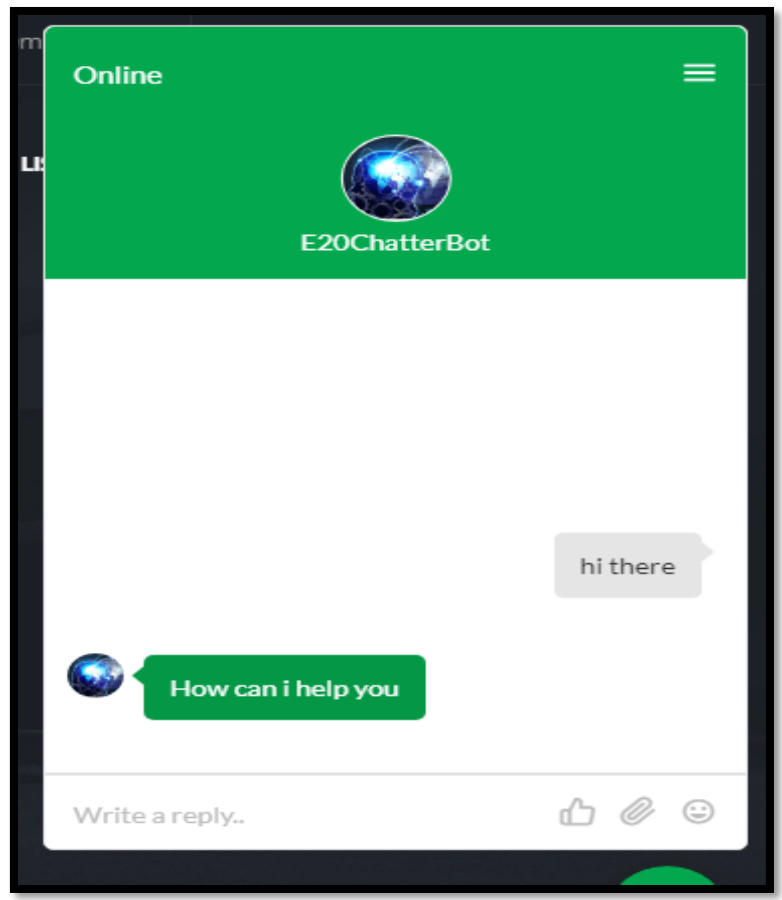

*Figure 37: Conversation with the Chatbot*

## 6.9 TESTING

The testing phase of SDLC is one the significant part of the system development lifecycle, as it shows the performance of the implemented system. It also tells the certain limitations and the extent of success of the system. At this stage of SDLC, various bugs and errors are checked, which alters the overall performance of the system, and they are recorded for further improvement. So, this feature can ensure the efficient performance of the chatbot system. We have performed a variety of tests in multiple ways to check the effectiveness of the implemented system.

While considering the testing phase of our SDLC of the chatbot system, the following are some of the tests we have done to check the performance of our system:

- White Box testing
- Black Box testing
- Stress testing

## 6.9.1 WHITE BOX TESTING

White box testing is one of the effective means of testing any implemented system based on programming. It is also called as glass-box testing in other words.

As the implementation phase is completed, now we are considering software for the testing purpose. It will ensure the operation of the implemented system. There are several ways that can be studied to test the application code to figure out if they are working accordingly. We have used a couple of software to analyse coding, and at certain places, we have discovered errors; however, we debug them.

We have coded the chatterbot software by pycharm; when we run the program, we got a couple of errors as the screenshot shown below:

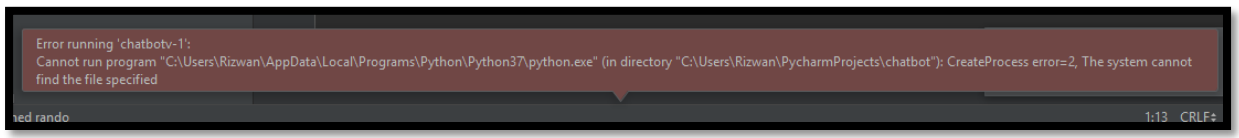

*Figure 38: Errors*

After debugging, our pycharm showed the results without error. We have done troubleshooting by a various online medium, and ultimately software started working fine. Below screenshot shows the result of white box testing after debugging.

```python
from chatterbot.trainers import ListTrainer # method to train the chatbot
from chatterbot import ChatBot # import the chatbot

bot = ChatBot('Test') # create the chatbot

conv = open('chats.txt', 'r').readlines()

bot.set_trainer(ListTrainer) # set the trainer

bot.train(conv) # train the bot

while True:
    request = input('You: ')
    response = bot.get_response(request)

    print('Bot: ', response)
```

*Figure 39: After Debugging*

## 6.9.2 BLACK BOX TESTING

Black box testing is primarily concerned with the behaviour of the system. As this testing system is followed by the input and output values, therefore the information available about the system becomes limited. If there are issues or errors in the algorithm, this test will display them. To check the functionality of the system, we carry out tests on an individual task basis, in the way that each test is performed with a single function individually. Various high-level function testing has a comparison with domain-testing.

The functional testing of our project shows the intended operation of our chatbot system. It can be seen from the picture below that when the user launched the website, the chatbot automatically pops up, asking the user for help and conversation can be continued. This test proves the normal function of the implemented chatbot system.

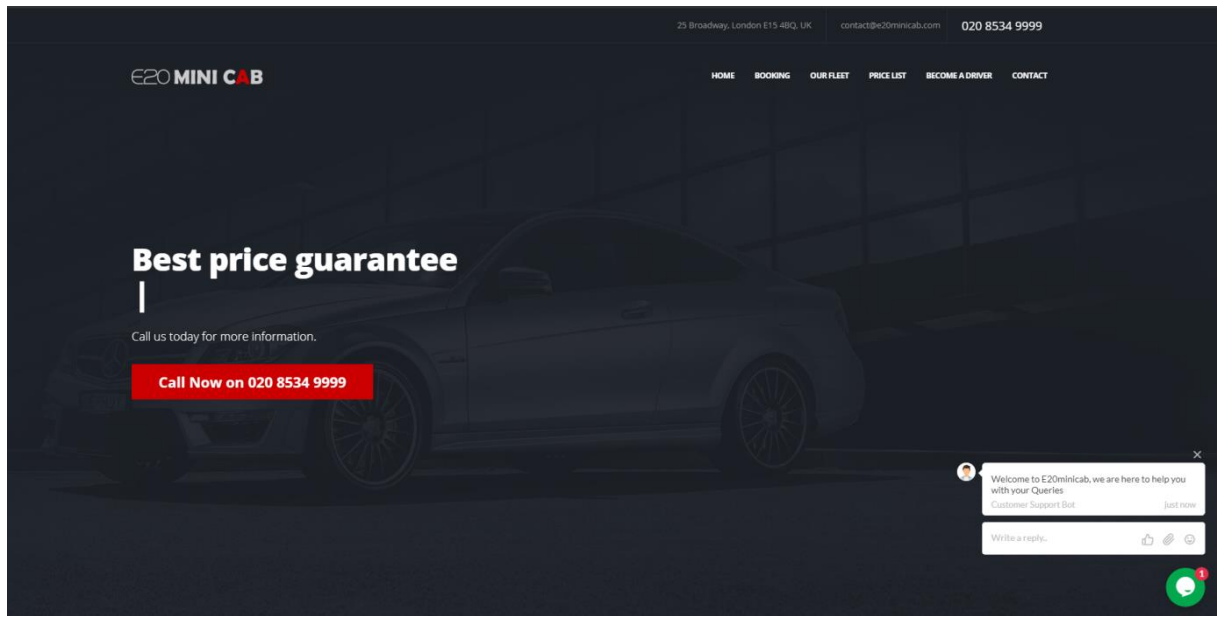

*Figure 40: Chatbot with the Web interface*

After this, we have done other functional testing to find out the operation of chatbot system in terms of conversation. \below screenshot shows the respond of the system based on the user input and this conversation can be continued. So, this test proves the functional testing of the system in terms of giving replies to the user.

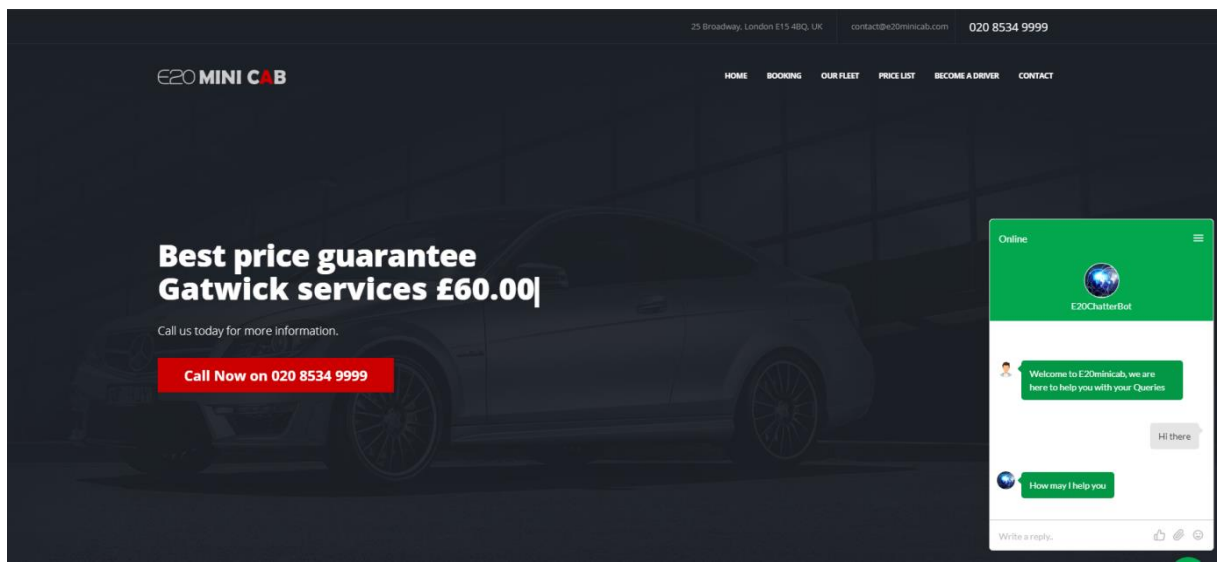

*Figure 41: Functionality of chatbot*

The screenshot below shows that the chatbot does not understand the questions or the queries that are not related to the company and which it has not learned.

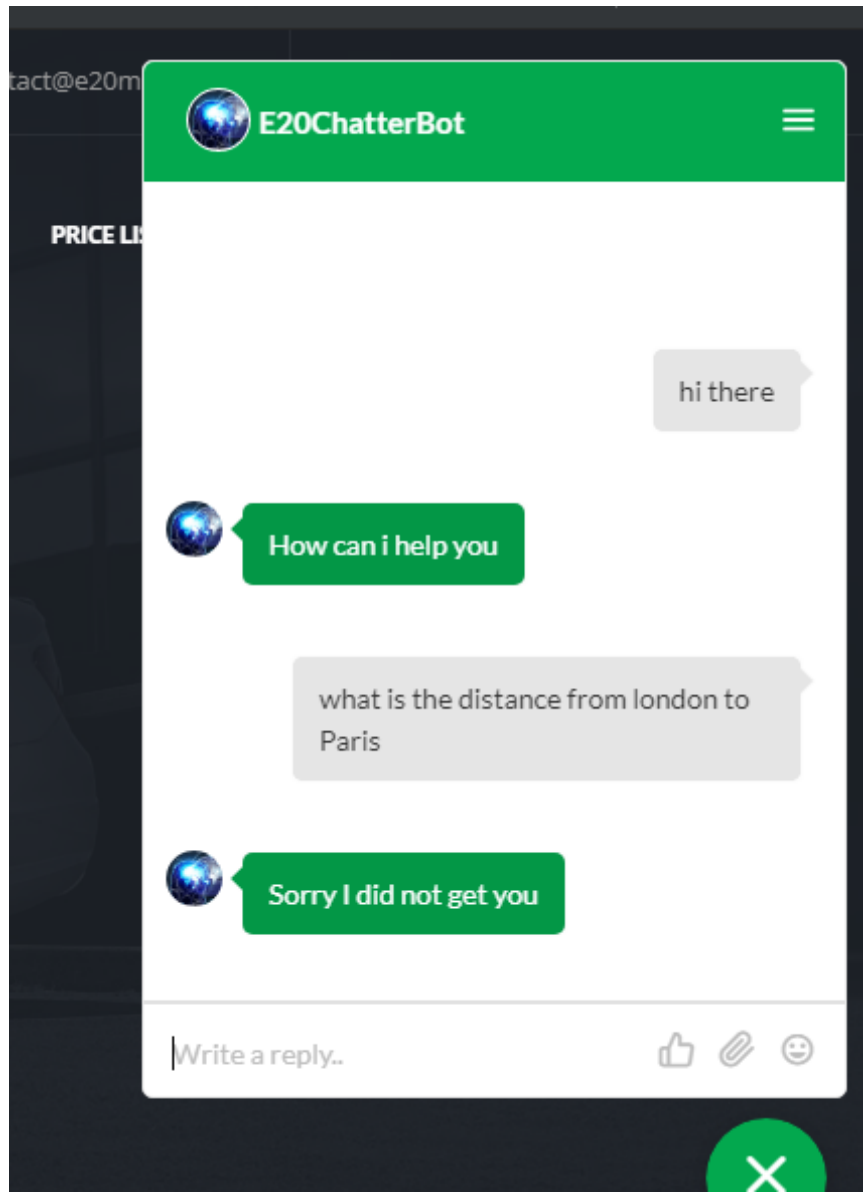

*Figure 42: Response to the Unknown Query*

## 6.9.4 STRESS TESTING

This type of testing is concerned with the alternation of the program out of its limits, in the way that system behaviour is monitored in extreme circumstances (stress). There is a probability of seeing the system not operating as usual, and this can be useful to see the behaviour of the system in case of any issue or error. The understanding of the behaviour trend can be useful to check the vulnerability of the system during extreme cases, and thus various tests techniques can be applied to reduce the risk of potential issues. It required a right hand on approach on the troubleshooting of the system to find error resolution.

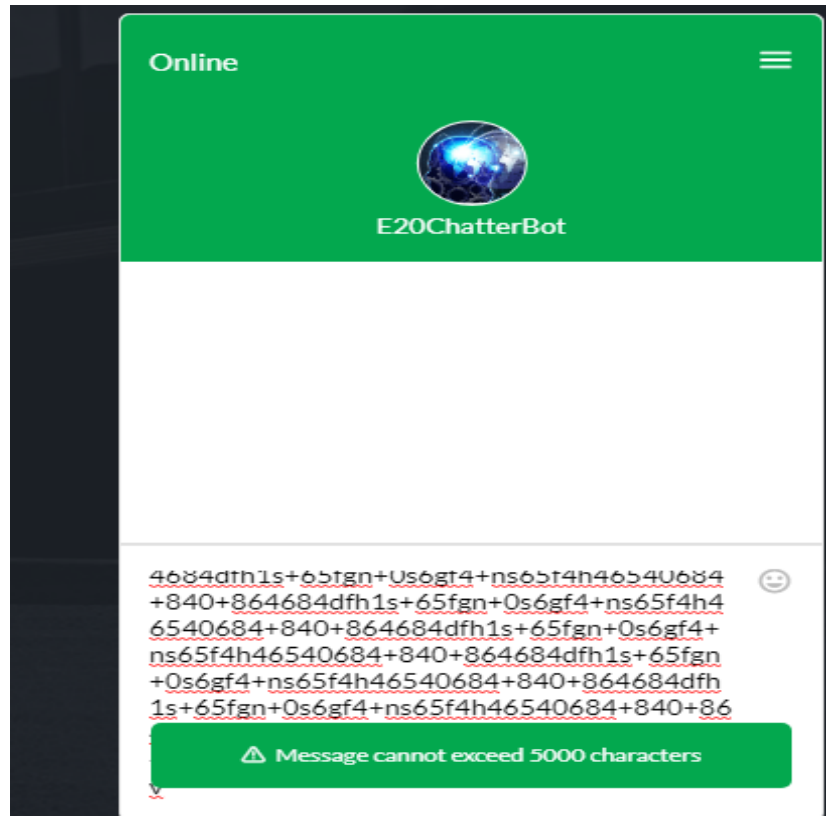

*Figure 43: Stress Testing*

In our project, we have performed stress testing by putting over 5000 characters, and it was actually to test the capability to check if the system can support a certain number of characters. So, after testing the above picture shows the message once it reaches to 5000 characters. It can be concluded from this test result, that this system cannot support characters over 5000, and it also disable the "Enter" key button.

## 6.10 ERROR HANDLING

In the development of the software system, we could encounter a variety of errors. Following are some of the types of errors:

- The error caused by the syntax of the programming language.
- The semantic error can be caused due to a lack of efficiency in the use of program statement.
- The logical error can be caused if the program statement is not satisfied. Although the program is executed with no error, however, the expected outcome is not generated.

On the basis of the errors mentioned above, we have distinguished:

Compile-time error: Semantic and Syntax error is distinguished by the compiler.

Runtime error: It includes semantic (dynamic) and logical error that is not distinguished by the compiler

```
Error running 'Chatterbot-v2':
Cannot run program "C:\Users\Rizwan\AppData\Local\Programs\Python\Python37\python.exe" (in directory "C:\Users\Rizwan\PycharmProjects\chatbot"): CreateProcess error=2, The system cannot find the file specified
```

After the integration of the chatterbot to the website we have found that the chatbot was not responding to the questions fast enough, so we are done troubleshooting and found that the "response_time" parameter was high and that was the reason, it was giving such a long delay to the response.

There was also an incident that chatterbot was not answering as expected, and we were giving a random response from the data list, we have found that the 'chatterbot.logic.BestMatch" was misconfigured and then we corrected the system.

# 7 CHAPTER

# CONCLUSION AND FUTURE WORK

## INFORMATION IN THIS CHAPTER

- ❖ OVERVIEW
- ❖ ANY BUSINESS COMMUNICATION IN THE LAST 12 MONTHS?
- ❖ ANY ONLINE BASED SERVICES OVER THE 12 LAST MONTH?
- ❖ WHAT IS THE FUTURE OF CHATBOT?
- ❖ ADVANTAGES OF THE CHATBOT SYSTEM?
- ❖ WHY WOULD YOU NOT CONSIDER CHATBOT SYSTEM?
- ❖ CHATBOT VS EMAIL
- ❖ CHATBOT VS PHONE
- ❖ CHATBOT VS APP

# 7.1 CONCLUSION OVERVIEW

The principal objective of this project work for the company was to build a chatbot system that is capable of generating answers to the complex questions by the customers. The consideration was to give well defined and logical replies. We have discussed tools and techniques that can be used for the implementation of the chatbot system and the selection of the best techniques that could fulfil the company requirements. The primary concern was the NLP, vector space model and machine learning model. Following are the deliverables of this project:

- The successful operation of the chatbot system
- Effective use of machine-learning techniques in the chatbot system
- Evaluation of the chosen technology

Following criteria are successfully achieved, yet the answering quality of the chatbot is not very high. Following are some of the reasons for low-quality answers of the chatbot system implemented on the company website:

There is a gap coming in the machine-learning model on q2q-datasets which we have used in the implementation of the chatbot. This is due to the smaller datasets of the questions and answers, which does not follow up with all the answers to the variety of questions. The result sometimes is unexpected when a user input question which is not known byte dataset used. On the other hand, if the question that matches with our datasets follows up with the justified and well-defined answer by the chatbot. It is quite logical if more than half of the user input questions does not synchronise with our chatbot data sets, it will give the worse result. Another justification for less defined answers is the specific data sets of individual questions. As the user input question is defined only to the inevitable question, therefore sometimes it conflicts with the new user input.

# 7.2 LIMITATIONS

There are some limitations discovered during the phase of testing, which is based on the fact of not retrieving the right answers concerning the user input question. We have identified logs indicating this issue. A user expects the higher efficiency in terms of getting satisfying answers to their questions which could include multiple topics and areas; however, this variety of information may not be available in the database. The implemented chatbot does not have a memory for storage in the way that if the user asks similar questions multiple times, the answer remains the same. We have not used Anaphora at this stage, which could be useful for the user, and the system is capable of referring to the previous question. It requires memory for its operation.

Another limitation is the lack of its capability to identify spelling checks, where the user has to correct the spelling and input the question again, manually. The system does not work with the errors of noun and verb in the sentences, so it is required for the users to follow the correct pattern, and if it is submitted in a wrong order, the user will have to submit the question again with the correct pattern of noun and verb in the sentence. For our project, the chatbot system is indeed useful for the company; however, it still requires improvement in its user interface to make it more user-friendly and efficient.

## 7.3 FUTURE WORK

Based on our research and the implementation of the chatbot system for the company, we have discovered following future work which can cause a significant improvement in this system:

- Firstly, our project covers limited question answers datasets comprise of smaller domains. We have considered around 150 questions answers, and we have realised that this number is not sufficient for the (customer) more extensive scale. This can be improved by extended datasets in the smaller domains.
- This chatbot is limited to follow the answer only to the specific answer, so the issue arises when some new question is input in a similar category. So, for future improvement, it is worth to re-write answers such that it can be used for multiple questions.
- This chatbot system is more specific to give an answer to the questions rather than starting a conversation with the users by providing information. So, this can be improved in the future by integrating more functionalities in the chatbot system that could be capable of making conversation rather than simply reply to a specific question.
- For the chatbot system that is based on the machine learning techniques, it could be beneficial to add hand-coded rules to answer matching questions, i.e. "what is the distance of Heathrow from London?". This can be quite useful, having less complicated questions.

- The general operation of the chatbot system for machine learning usually takes 15 seconds; however, the similar chatbot used by Scala takes a mini second to process enquiry. For future improvement, it can be beneficial to run the chatbot in Python web-based program with the machine learning techniques.
- As mentioned in the literature review about the advantages of contemplating the combined chatbot system that could use both NLP and vector-space model. This requires a significant amount of training data, however in the future is worth to consider the implementation of this chatbot system by the combination of these models.

The research that has been done the Machine Learning approach is very time-consuming. Our success on this project helps us understand that this complicated question and answer retrieval could be handled by a chatbot.

# 8. LIST OF REFERENCES

# 9. APPENDIX

## APPENDIXA [Interview snapshot]

Rebecca H: Hello LESLIE, my name is Rebecca, and welcome to E20minicabs!

LESLIE: hi

LESLIE: are you there? I want to know if I can return back from heathrow ?

Rebecca H: You may absolutely return, and the cost will be 65 pounds.

Rebecca H: what time you chose to return and what the destination ?

LESLIE: thank you, I would like to return at 6 pm and my destination is croydon.

Rebecca H: you can book the taxi now if you want over phone 028064565 .

LESLIE: Ok that's good. Thanks for your help.

Rebecca H: You're very welcome! Hope you have a great day!

LESLIE: Ok - thanks for your help.

Rebecca H: You're very welcome!

## APPENDIX B [Interview Snapshot]

Masud Rahi: Hello MAHMUD, my name is Masud, and welcome to E20minicabs!

MAHMUD: hi , how are you? I want to know what is the return ticket if I book a cab from bow to luton airport ?

Masud Rahi: the cost will be 55 pounds for one way and both ways is 100 pounds

MAHMUD: thank you, I would like to return at 6 pm .

Masud Rahi: you can book the taxi now if you want over phone 028064565 .

MAHMUD: Ok that's good. Thanks for your help.

Masud Rahi: You're very welcome! Hope you have a great day!

MAHMUD: Ok - thanks for your help.

Masud Rahi: You're very welcome!

# APPENDIX C [Suggestions from Supervisor]

**Paul Sant**
Mon 11/11/2019 10:28

Dear Rejwan,

Many thanks for sending this over. There are a few areas of imporvement

1) Firstly, the document needs to be professionally proof read as there are many grammatical mistakes which detract from the good design work that you have completed.

2) I would suggest you separate out the methodology (the setps undertaken) from the actual design - you can do this by clearly separating out 1) Methodology and 2) Design into separate sections.

3) Where you describe approaches for data collection (interviews etc.) be clearer around the fact that you used a survey based approach

It is good to see that you have included flowcharts, but these lack a numerical flow, or an ordering (e.g. top of the page to the bottom) and it is not immediately clear at to what order the steps are completed. Please add labels to indicate this otherwise the reader will not be able to follow.

You may also want to include some high level design of the chatbot components and how they interct as part of your design section.

By making these improvements you will be able to better explain the approach that you have taken.

I hope that helps,

Paul.

Dear Rejwan,

Please find attached some suggested revisions for your literature review chapter. I think the critical analysis needs to be more detailed and so we need to discuss what you are trying to convey to the reader.

Regards,

Chapter - 9: Appendix    III# APPENDIX D [My response to Supervisors email]

During the whole process of the placement, we had number of conversation and meeting with the supervisor. Because of the supervisor's guidance and suggestions I was on the right track of the project to accomplish in time.

1.

Rejwan Bin Sulaiman
Tue 12/11/2019 03:32

Good morning Dr Paul,
Thank you for your suggestions and advice. I have taken the following measures.

- I have bought proofreading software that will help me eliminate the mistakes
- I have separated the chapters into two parts, "METHODOLOGY" AND "DESIGN".
- I have taken interview and survey-based approach, I have written a paragraph where I mentioned, what are the procedures I have chosen for the project.
- I have added labels on graphs and flowcharts. Also, adjust with the numerical flow.
- I am also adding a high-level design of the chatbot components.

Thank you again for your time

2.

Good morning sir,
Thank you for your corrections and suggestions on the LR part of my thesis. I have amended the LR part accordingly. you have mentioned "*I think the critical analysis needs to be more detailed*", can you please elaborate what details will I add. You can give hints with questions and I will try to provide explanations based on those questions for more detailed and perfect "critical evaluation".

I have started writing my Methodology part. Please find attached, a portion from methodology part. In term of data collection, do you suggest to use screenshots of the questions and survey in the main body or you suggest to use in the Appendix?

Expecting your suggestions and advice.
Thank you.

Regards:

III | P a g e

# APPENDIX E [Website code files ]

| Name | Da |
|---|---|
| 📁 admin | 04 |
| 📁 assets | 22 |
| 📁 includes | 04 |
| car-listing.php | 04 |
| carrental.zip | 11 |
| check_availability.php | 18 |
| contact-us.php | 26 |
| index.php | 04 |
| logout.php | 14 |
| my-booking.php | 27 |
| my-testimonials.php | 27 |
| page.php | 27 |
| post-testimonial.php | 27 |
| profile.php | 27 |
| search-carresult.php | 25 |
| update-password.php | 27 |
| vehical-details.php | 04 |

# APPENDIX F [Chatbot code that did not work]

```python
import requests
import string
from lxml import html
from googlesearch import search
from bs4 import BeautifulSoup

# to search
# print(chatbot_query('how old is samuel l jackson'))

def chatbot_query(query, index=0):
    fallback = 'Sorry, I cannot think of a reply for that.'
    result = ''

    try:
        search_result_list = list(search(query, tld="co.in", num=10, stop=3, pause=1))

        page = requests.get(search_result_list[index])

        tree = html.fromstring(page.content)

        soup = BeautifulSoup(page.content, features="lxml")

        article_text = ''
        article = soup.findAll('p')
        for element in article:
            article_text += '\n' + ''.join(element.findAll(text = True))
        article_text = article_text.replace('\n', '')
        first_sentence = article_text.split('.')
        first_sentence = first_sentence[0].split('?')[0]

        chars_without_whitespace = first_sentence.translate(
            { ord(c): None for c in string.whitespace }
        )

        if len(chars_without_whitespace) > 0:
```

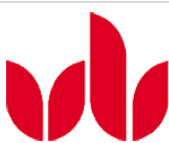

```
25          for element in article:
26              article_text += '\n' + ''.join(element.findAll(text = True))
27          article_text = article_text.replace('\n', '')
28          first_sentence = article_text.split('.')
29          first_sentence = first_sentence[0].split('?')[0]
30
31          chars_without_whitespace = first_sentence.translate(
32              { ord(c): None for c in string.whitespace }
33          )
34
35          if len(chars_without_whitespace) > 0:
36              result = first_sentence
37          else:
38              result = fallback
39
40          return result
41      except:
42          if len(result) == 0: result = fallback
43          return result
```

# APPENDIX G [Chatbot code did not worked for us]

```
    ["Nagesh created me using Python's NLTK library ","top secret
;)",]
    ],
    [
        r"(.*) (location|city) ?",
        ['Chennai, Tamil Nadu',]
    ],
    [
        r"how is weather in (.*)?",
        ["Weather in %1 is awesome like always","Too hot man here in
%1","Too cold man here in %1","Never even heard about %1"]
    ],
    [
        r"i work in (.*)?",
        ["%1 is an Amazing company, I have heard about it. But they
are in huge loss these days.",]
    ],

[
        r"(.*)raining in (.*)",
        ["No rain since last week here in %2","Damn its raining too
much here in %2"]
    ],
    [
        r"how (.*) health(.*)",
        ["I'm a computer program, so I'm always healthy ",]
    ],
    [
        r"(.*) (sports|game) ?",
        ["I'm a very big fan of Football",]
    ],
    [
        r"who (.*) sportsperson ?",
        ["Messy","Ronaldo","Roony"]

],
    [
        r"who (.*) (moviestar|actor)?",
        ["Brad Pitt"]

],
    [
        r"quit",
        ["BBye take care. See you soon :) ","It was nice talking to
you. See you soon :)"]
```

```python
from nltk.chat.util import Chat, reflections

pairs = [
    [
        r"my name is (.*)",
        ["Hello %1, How are you today ?",]
    ],
    [
        r"what is your name ?",
        ["My name is Chatty and I'm a chatbot ?",]
    ],
    [
        r"how are you ?",
        ["I'm doing good\nHow about You ?",]
    ],
    [
        r"sorry (.*)",
        ["Its alright","Its OK, never mind",]
    ],
    [
        r"i'm (.*) doing good",
        ["Nice to hear that","Alright :)",]
    ],
    [
        r"hi|hey|hello",
        ["Hello", "Hey there",]
    ],
    [
        r"(.*) age?",
        ["I'm a computer program dude\nSeriously you are asking me this?",]
    ],
    [
        r"what (.*) want ?",
        ["Make me an offer I can't refuse",]
```

## APPENDIX H [Chatbot did not work]

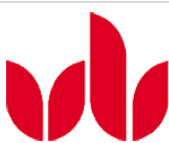

```python
#import necessary libraries
import io
import random
import string # to process standard python strings
import warnings
import numpy as np
from sklearn.feature_extraction.text import TfidfVectorizer
from sklearn.metrics.pairwise import cosine_similarity
import warnings
warnings.filterwarnings('ignore')

import nltk
from nltk.stem import WordNetLemmatizer
nltk.download('popular', quiet=True) # for downloading packages

# uncomment the following only the first time
#nltk.download('punkt') # first-time use only
#nltk.download('wordnet') # first-time use only

#Reading in the corpus
with open('chatbot.txt','r', encoding='utf8', errors ='ignore') as fin:
    raw = fin.read().lower()

#TOkenisation
sent_tokens = nltk.sent_tokenize(raw)# converts to list of sentences
word_tokens = nltk.word_tokenize(raw)# converts to list of words

# Preprocessing
lemmer = WordNetLemmatizer()
```

```python
def LemTokens(tokens):
    return [lemmer.lemmatize(token) for token in tokens]
remove_punct_dict = dict((ord(punct), None) for punct in string.punctuation)
def LemNormalize(text):
    return LemTokens(nltk.word_tokenize(text.lower().translate(remove_punct_dict)))

# Keyword Matching
GREETING_INPUTS = ("hello", "hi", "greetings", "sup", "what's up","hey",)
GREETING_RESPONSES = ["hi", "hey", "*nods*", "hi there", "hello", "I am glad! You are talking to me"]

def greeting(sentence):
    """If user's input is a greeting, return a greeting response"""
    for word in sentence.split():
        if word.lower() in GREETING_INPUTS:
            return random.choice(GREETING_RESPONSES)

# Generating response
def response(user_response):
    robo_response=''
    sent_tokens.append(user_response)
    TfidfVec = TfidfVectorizer(tokenizer=LemNormalize, stop_words='english')
    tfidf = TfidfVec.fit_transform(sent_tokens)
    vals = cosine_similarity(tfidf[-1], tfidf)
    idx=vals.argsort()[0][-2]
    flat = vals.flatten()
    flat.sort()
    req_tfidf = flat[-2]
    if(req_tfidf==0):

        robo_response=robo_response+"I am sorry! I don't understand you"
        return robo_response
    else:
        robo_response = robo_response+sent_tokens[idx]
        return robo_response

flag=True
print("ROBO: My name is Robo. I will answer your queries about Chatbots. If you want to exit, type Bye!")
while(flag==True):
    user_response = input()
    user_response=user_response.lower()
    if(user_response!='bye'):
        if(user_response=='thanks' or user_response=='thank you' ):
            flag=False
            print("ROBO: You are welcome..")
        else:
            if(greeting(user_response)!=None):
                print("ROBO: "+greeting(user_response))
            else:
                print("ROBO: ",end="")
                print(response(user_response))
                sent_tokens.remove(user_response)
    else:
        flag=False
        print("ROBO: Bye! take care..")
```

# APPENDIX I [Ethical consideration]

**SECTION B    Check List**

Please answer the following questions by circling **YES** or **NO** as appropriate.

1. Does the study involve vulnerable participants or those unable to give informed consent (e.g. children, people with learning disabilities, your own students)?

   YES        (NO)

2. Will the study require permission of a gatekeeper for access to participants (e.g. schools, self-help groups, residential homes)?

   YES        (NO)

3. Will it be necessary for participants to be involved without consent (e.g. covert observation in non-public places)?

   YES        (NO)

4. Will the study involve sensitive topics (e.g. obtaining information about sexual activity, substance abuse)?

   YES        (NO)

5. Will blood, tissue samples or any other substances be taken from participants?

   YES        (NO)

6. Will the research involve intrusive interventions (e.g. the administration of drugs, hypnosis, physical exercise)?

   YES        (NO)

7. Will financial or other inducements be offered to participants (except reasonable expenses or small tokens of appreciation)?

   YES        (NO)

8. Will the research investigate any aspect of illegal activity (e.g. drugs, crime, underage alcohol consumption or sexual activity)?

   YES        (NO)

9. Will participants be stressed beyond what is considered normal for them?

   YES        (NO)

10. Will the study involve participants from the NHS (patients or staff) or will data be obtained from NHS premises?

    YES        (NO)

*If the answer to any of the questions above is "Yes", or if there are any other significant ethical issues,*

# APPENDIX J [SDLC CONSIDERATION]

It is necessary to have a discussion on the Research Onion before going into the details.

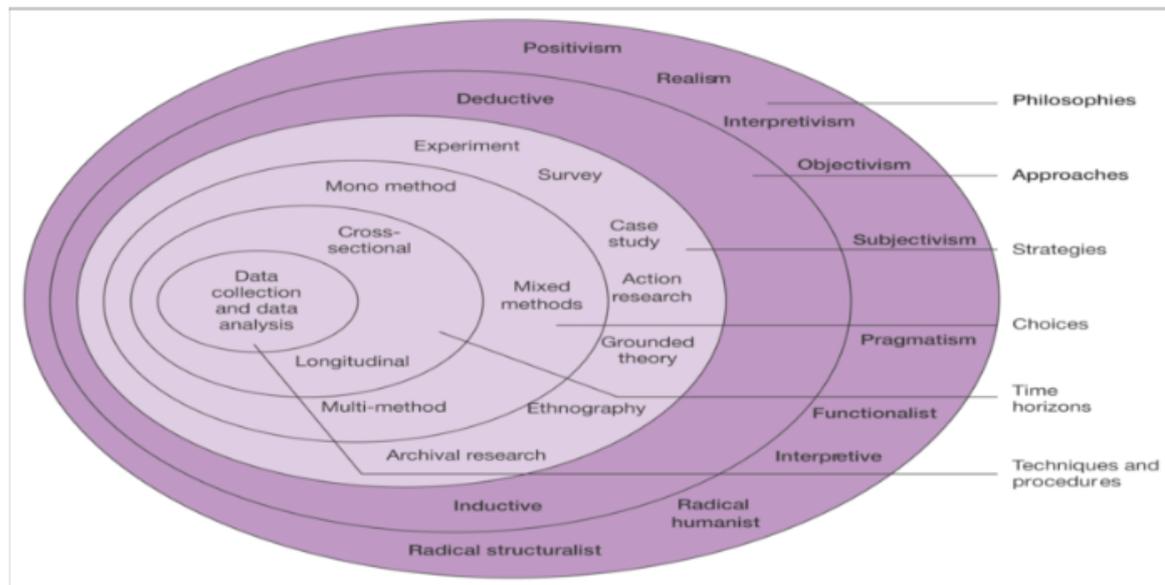

The research onion portrays the step-by-step process of the research which entails the activities the research does and the views that the researcher holds. It is mandatory to go for in-depth analysis and the layers of the onion should be done thoroughly up until the desired outcome of the research.

Saunders (2009) had come up with four philosophies on the research which would let the researchers utilize one philosophy that is the amalgamation of two philosophies technically. *Pragmatic* research philosophy depicts the research process in a multi-dimensional way and interprets the process of that research. In addition, the data have the inclusion of two important paradigms which are called interpretivism and positivism (Saunders , et al., 2009). The positivist research paradigm relies on the collection of data whereas the interpretation focuses on the source of the gathered data. In addition, interpretivism research paradigm can pave way to analysis further so that a possible solution can be achieved (Saunders , et al., 2009).